\documentclass{bmvc2k}

\title{Beyond Parameter Space: NTK-Guided
Personalized Aggregation for Robust
Federated Learning}

\addauthor{Mirko Konstantin}{konstantin@zib.de}{1,2}
\addauthor{Stefan Zachow}{zachow@zib.de}{1}
\addauthor{Anirban Mukhopadhyay}{anirban.mukhopadhyay@tu-darmstadt.de}{2}

\addinstitution{
 Zuse Institute Berlin (ZIB)\\
 Takustraße 7\\
 14195 Berlin, Germany
}
\addinstitution{
 Technical University of Darmstadt\\
 Karolinenplatz 5\\
 64289 Darmstadt, Germany
}

\runninghead{Konstantin et al.}{NTK-Guided
Personalized and Robust Aggregation}

\usepackage{booktabs} 
\usepackage{amsmath}
\usepackage{amssymb}
\begin{document}

\maketitle

\begin{abstract}
Federated learning (FL) enables collaborative model training across distributed clients while preserving data privacy by keeping data local. A central challenge in FL is determining which client updates are beneficial for aggregation with respect to each client’s target domain. Existing methods typically address this problem in parameter space by comparing model parameters or gradients. However, parameter-space similarity is often a poor proxy for predictive behavior, particularly in heterogeneous settings where client data is not independent and identically distributed. As a result, updates that are misaligned with a client’s target domain, including those arising from heterogeneous data distributions or malfunctioning clients, may be incorporated into aggregation and degrade local model performance.
In this work, we propose \underline{\textbf{L}}ocal \underline{\textbf{I}}nference \underline{\textbf{G}}uided Aggregation for \underline{\textbf{H}}eterogeneous \underline{\textbf{T}}raining Environments to \underline{\textbf{Y}}ield \underline{\textbf{E}}nhancement Through \underline{\textbf{A}}greement and \underline{\textbf{R}}egularization (\textbf{LIGHTYEAR}), a federated learning framework that performs update selection in the function space. Central to our method is an NTK-based agreement score that characterizes predictive behavior and is used to determine the optimal aggregation set for each client. By relating model parameters to local predictive responses, the Neural Tangent Kernel (NTK) enables a function-space characterization of model updates and provides a more expressive criterion for update selection than parameter-space similarity alone.
Since access to function-space information before aggregation is not available in conventional centralized FL, LIGHTYEAR leverages a peer-to-peer (P2P) topology, where clients exchange updates directly and can locally evaluate incoming models on private validation data. This enables each client to construct a personalized aggregation set consisting only of updates that are beneficial with respect to its own target domain. The selected updates are then aggregated using a regularized aggregation rule that further stabilizes training under heterogeneity.
Our empirical evaluation across five datasets and nine baseline methods demonstrates that LIGHTYEAR consistently outperforms both centralized FL baselines and existing P2P approaches. \textbf{The code is available at:} \url{https://github.com/MECLabTUDA/LIGHTYEAR}

\end{abstract}
\section{Introduction}
\label{sec:intro}

In recent years, federated learning (FL) has emerged as a powerful approach for collaboratively training machine learning models across a network of distributed clients, without requiring the exchange of raw data \cite{mcmahan2017communication}. This paradigm has gained increasing attention due to its natural fit for privacy-sensitive applications such as healthcare, mobile personalization, and finance \cite{taiello2024enhancing,ali2022federated,lemke2025equitable}. FL was originally introduced in a centralized star-shaped topology, where a central server coordinates communication and aggregates updates received from participating clients \cite{mcmahan2017communication}. While this formulation is conceptually simple and effective in homogeneous settings, it faces substantial limitations under real-world conditions characterized by heterogeneous data distributions and unreliable clients \cite{zhao2018federated}.

A central objective in FL is to learn models that generalize well to the target domain of each participating client. In practice, however, performance on a client’s target domain can degrade due to two major factors. First, clients frequently operate on non-independent and identically distributed (non-IID) data. Differences in acquisition protocols, sensing conditions, hardware configurations, or user behavior result in client-specific data distributions, violating exchangeability of data samples across the federation \cite{zhu2021federated,islam2024fedclust}. Consequently, updates learned by one client are not equally beneficial for all others. Second, client reliability is inherently variable, and malfunctioning clients may contribute corrupted updates due to faulty sensors, broken data pipelines, or malicious attacks such as model poisoning \cite{konstantin2024asmr}. Both heterogeneity and malfunctioning updates can negatively affect model performance on the target domain of individual clients.
As a result, not every client benefits equally from aggregating updates from the full federation. Instead, each client benefits most from a personalized subset of client updates that best aligns with its own target domain and learning objective \cite{chen2024personalized}. This motivates the need for a personalized client-specific solution, where each client selectively incorporates only the most beneficial updates while excluding irrelevant or harmful ones.
\begin{figure}
\centering
\includegraphics[width=\linewidth]{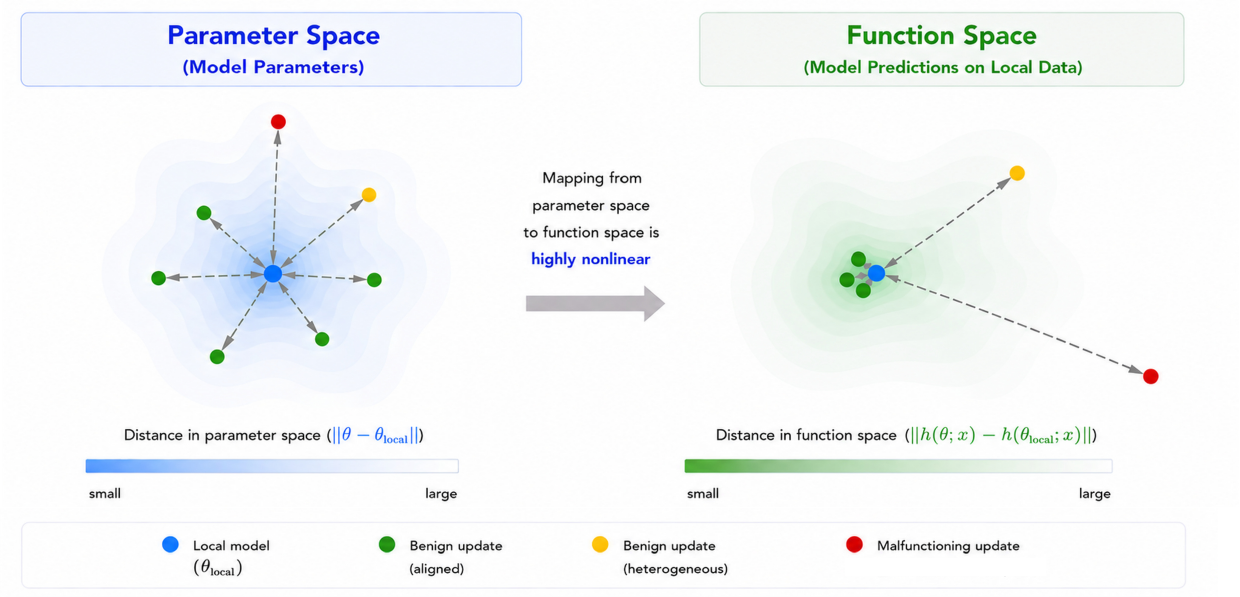}
\caption{Illustration of the discrepancy between parameter space and function space. While client updates may appear similarly close in parameter space, their effect on model predictions can differ substantially in function space. This illustrates that parameter distance alone is an unreliable proxy for model behavior under heterogeneous settings.}
\label{fig:motivation}
\end{figure}

Traditional FL methods attempt to identify beneficial updates through operations in parameter space, typically by comparing model parameters or gradients \cite{blanchard2017machine,sattler2020byzantine,xu2022byzantine,pillutla2022robust,munoz2019byzantine}. This is primarily a consequence of centralized FL itself, where the server has access only to model updates, but not to the underlying client data distributions. Without access to data distribution or training behavior, the server cannot directly assess how an incoming update behaves on a client’s target domain and therefore must rely on parameter-space proxies. However, parameter space is often insufficient for this task. Prior work has shown that distances in parameter space do not consistently correlate with similarity in predictive behavior \cite{benjamin2018measuring}. Models with small parameter differences may lead to substantially different behavior, which is illustrated in Figure \ref{fig:motivation}. Therefore, parameter space can be a poor proxy for actual model behavior.
In contrast, evaluating models in function space provides a more expressive representation of how a model behaves on a given data distribution \cite{benjamin2018measuring}. Function-space information captures predictive behavior directly and therefore provides a more meaningful basis for determining whether an update is beneficial for aggregation with respect to a client’s target domain. The challenge, however, is that this information is inaccessible in centralized FL before aggregation, since the central server has no access to local data \cite{liu2022threats} and cannot evaluate incoming models on client-specific validation distributions.

To address this limitation, we leverage a decentralized peer-to-peer (P2P) training topology \cite{gabrielli2023survey,warnat2021swarm}. By moving away from the centralized star topology, clients exchange updates directly without relying on a central server. This communication structure provides each client with access to neighboring updates before aggregation and enables local evaluation on private validation data. As a result, every client can determine the predictive behavior of incoming models with respect to its own local target distribution and construct a personalized aggregation set based on function-space behavior rather than parameter similarity.
Based on this idea, we introduce the \underline{\textbf{L}}ocal \underline{\textbf{I}}nference \underline{\textbf{G}}uided Aggregation for \underline{\textbf{H}}eterogeneous \underline{\textbf{T}}raining Environments to \underline{\textbf{Y}}ield \underline{\textbf{E}}nhancement Through \underline{\textbf{A}}greement and \underline{\textbf{R}}egularization \textbf{(LIGHTYEAR)}. LIGHTYEAR is a P2P FL framework designed to select a personalized aggregation set for each client by measuring alignment in function space. To estimate model behavior, LIGHTYEAR leverages the Neural Tangent Kernel (NTK) \cite{jacot2018neural}, which relates parameter changes to predictive behavior and captures how a model responds locally on a given dataset. Using the NTK, we compute an agreement score that measures the similarity of predictive behavior between client models on a local validation set. This enables each client to identify which incoming updates are most beneficial with respect to its own target domain and selectively incorporate only those into aggregation. Finally, we aggregate the selected updates using a regularization term to mitigate client drift in heterogeneous environments.
As a result, LIGHTYEAR enables client-specific model aggregation with fine-grained control over update selection in the function space. This improves robustness against malfunctioning or irrelevant updates, reduces performance drops on the target domain, and enhances personalization in heterogeneous federated learning environments.\\
Our main contributions are as follows:

\textbf{We propose LIGHTYEAR}, a framework that combines a function-space agreement score for update selection with a regularized aggregation rule, enabled through P2P FL to support robust and personalized model training.

\textbf{We introduce the agreement score}, a metric based on the NTK that captures semantic alignment between client updates and the local model in function space by relating parameter changes to predictive behavior on the target domain. This score is used to select the personalized set of updates for aggregation.

\textbf{We propose a regularized aggregation rule}, which includes a round-dependent regularization term that controls the influence of updates over time, mitigating the effects of client drift in heterogeneous environments.


\section{Related Work}
\label{sec:relatedwork}

To address prediction errors in FL, two main research directions have emerged:
Robust FL focuses on mitigating the impact of malfunctioning clients and personalized FL, which tackles errors caused by distribution shifts. \\

\textbf{Robust Federated Learning}: Malfunctioning clients pose a significant threat to FL by submitting harmful updates, whether due to adversarial intent or technical faults, that degrade the performance of the aggregated model when incorporated into the global update \cite{zhang2022fldetector,konstantin2024asmr}. To mitigate the impact of such corrupted contributions, a variety of robust aggregation methods have been proposed. Originally introduced for centralized FL, these methods are typically performed at the server level, where no access to client-side data or reference performance metrics is available \cite{zhang2024anomaly}. As a result, these approaches rely solely on distance-based measures computed over model parameters or gradients to assess the similarity between updates. Clients whose updates deviate significantly from the majority are either down-weighted during aggregation \cite{pillutla2022robust,cao2020fltrust,mhamdi2018hidden} or excluded entirely \cite{blanchard2017machine,sattler2020byzantine,yin2018byzantine}. Alternative approaches, such as FedTrans \cite{yang2024fedtrans}, improve robustness through utility-based client selection using a shared server-side dataset. With the growing interest in decentralized FL, robust aggregation methods have been adapted accordingly. In this setting, robustness must be ensured at the client level, and distance-based mechanisms are employed locally to evaluate incoming updates \cite{fang2024byzantine,he2022byzantine}.\\

\textbf{Personalized Federated Learning}:
To address the limitations of the naive FedAvg approach in heterogeneous settings, where clients
operate on non-IID data, personalized federated learning (pFL) was introduced to better adapt the global models to local client requirements. Rather than relying solely on a single shared model, pFL methods aim to tailor the learned representations to each client’s domain. Ditto \cite{li2021ditto} extends this concept by maintaining two separate models per client: one global model for participation in communication rounds, and one personalized model for local inference, with training guided by a regularization term that encourages consistency between the two. FedALA \cite{zhang2023fedala} introduces an additional local aggregation step, where each client combines the global model and its locally trained model using a weighted average, thereby
adapting the global knowledge more flexibly to the local context.

\section{Problem Statement}
\label{sec:problem}
We consider a decentralized FL setting in which a set of clients $P = \{1, \ldots, N\}$
collaborate to learn a model. The network is structured as a P2P topology, where each client communicates directly with all other clients. Formally, for each client $i \in P$,
let $\mathcal{N}(i) \subseteq P \setminus \{i\}$ denote the set of neighboring clients from whom it receives model updates
$\{\theta_j\}_{j \in \mathcal{N}(i)}$, where $\theta_j$ denotes the parameters of the local model of client j. The client then aggregates these models with its own to update its parameters.
A central challenge in this setting is that client $i$ has no information about the training data, optimization procedure, or reliability of its neighbors' model updates $\{\theta_j\}_{j \in \mathcal{N}(i)}$. Consequently, it cannot determine a priori which of the updates will be beneficial or detrimental to its local objective. The naive aggregation of all received models,
\begin{equation}
    \theta_i^{\text{new}} \leftarrow \text{Aggregate} \left( \{\theta_j\}_{j \in \mathcal{N}(i)} \cup \{\theta_i\} \right),
\end{equation}
may introduce harmful biases.
In particular, blindly aggregating all updates can degrade model performance in the following ways: 

\textbf{Malicious clients} may submit adversarial updates $\theta_j^{\text{mal}}$ with the intent of corrupting the learning process.

\textbf{Unreliable clients} may unintentionally produce corrupted updates $\theta_j^{\text{faulty}}$, e.g., due to defective data pipelines, incorrect labels, or faulty image acquisition.

\textbf{Non-exchangeability across clients}: Even in benign scenarios, updates from clients trained on data distributions $\mathcal{D}_j$ significantly different from the client’s own distribution $\mathcal{D}_i$ may lead to performance degradation when aggregated \cite{zhao2018federated}.\\
\begin{figure}
    \centering
    \includegraphics[width=0.7\linewidth]{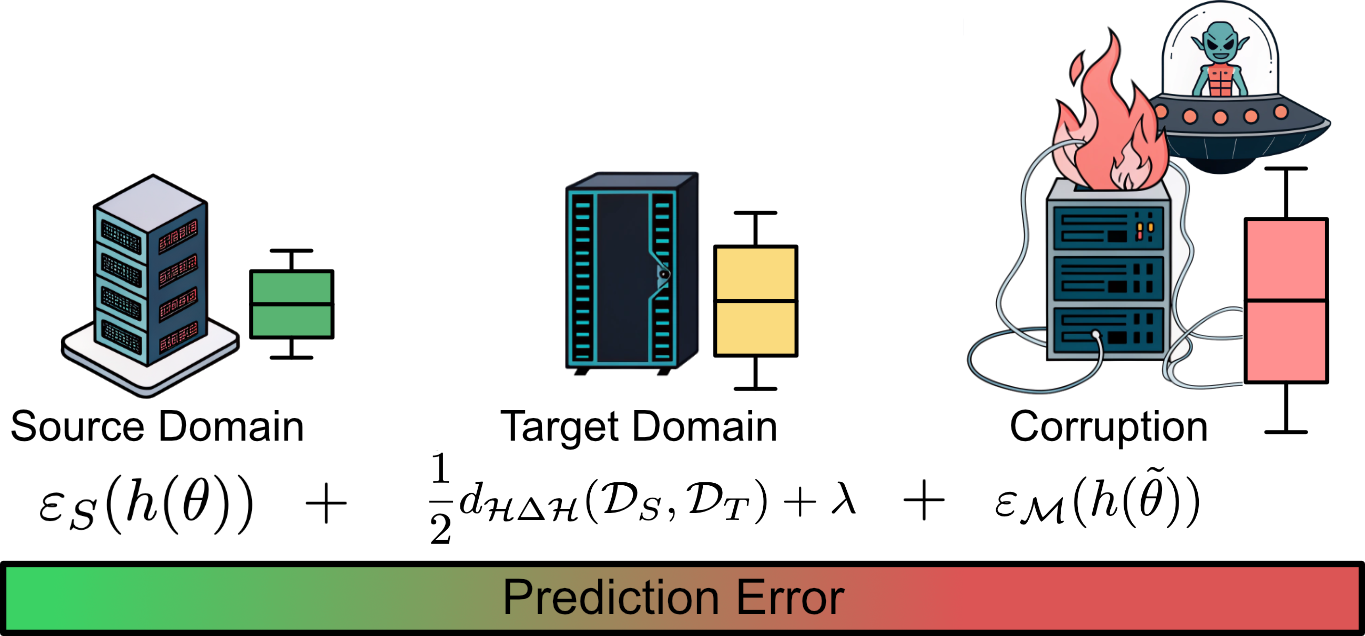}
    \caption{Illustration of the decomposition of the prediction error. The boxplot displays the error across the instances, with color indicating the magnitude of the error (green: low error, red: high error).}
    \label{fig:errors}
\end{figure}
In general, model updates from such clients often exhibit structural divergence from the models
trained on locally aligned data. The training data can be described as $<\mathcal{D}, f>$, where $\mathcal{D}$ is the distribution with input data $\mathcal{X}$ and $f$ is the labeling function
$f: \mathcal{X} \rightarrow [0, .., n_c]$, where $n_c$ is the number of classes. Let $\varepsilon_i(h(\theta))$ denote the prediction error of model $h$ with parameters $\theta$ on client $i$’s distribution $\mathcal{D}_i$, such that
\begin{equation}
\label{eq:domain_error}
    \varepsilon_i(h(\theta)) = \mathbb{E}_{\mathbf{x} \sim D_i} \left[ |h(\theta;\mathbf{x}) - f_i(\mathbf{x})| \right].
\end{equation}
The goal is to identify the aggregation set $\mathcal{S}_i \subseteq \mathcal{N}(i)$ of client updates such that:
\begin{equation}
    \theta_i^{\text{new}} \leftarrow \text{Aggregate} \left( \{\theta_j\}_{j \in \mathcal{S}_i} \cup \{\theta_i\} \right),
\end{equation}
where $ \varepsilon_i(h(\theta)_i^{\text{new}}) \text{is minimized}$.

\textbf{The fundamental problem is:}
\textit{Given a set of client updates $\{\theta_j\}_{j \in \mathcal{N}(i)}$ and no access to their underlying data distributions or training errors, determine an optimal aggregation subset $\mathcal{S}_i \subseteq \mathcal{N}(i)$ that maximizes local performance on $\mathcal{D}_i$.}
This requires a reliable indicator for distributional similarity or alignment between updates and the local data of client $i$, which is the core methodological focus of our work.

\section{Methodology}
\label{sec:methodology}
In the following, we begin by describing the prediction error a model incurs on a target distribution. This formulation accounts for both the violation of exchangeability between the source and target data distributions and the potential corruption of the model’s parameters due to malfunctions. We then highlight that P2P topologies offer a unique opportunity to approximate this error in a decentralized manner, in contrast to traditional star-shaped topologies. Finally, we introduce a novel agreement-based method that enables each client to select a custom aggregation set, thereby improving robustness and enhancing performance on the target distribution.

\subsection{Exchangeability in Federated Learning}
To consider the second case of disadvantageous clients, we must examine the concept of exchangeability, which plays a crucial role in the effectiveness of model aggregation. In statistical learning theory, a dataset $\{Z_i\}_{i=1}^{n}$ is said to be \textit{exchangeable} if its joint distribution is invariant under permutations.
Formally, a sequence $Z_1, \ldots, Z_n$ is exchangeable if for any permutation $\pi$ of $\{1, \ldots, n\}$, the joint distribution satisfies:
\begin{equation}
    P(Z_1, \ldots, Z_n) = P(Z_{\pi(1)}, \ldots, Z_{\pi(n)}).
\end{equation}
This condition generalizes the IID assumption and underpins the validity of many theoretical guarantees in machine learning, such as those provided by conformal prediction and generalization bounds.
In the context of FL, however, the assumption of data exchangeability across clients
is often violated. Each client $i \in P$ may possess data drawn from a distinct underlying distribution $\mathcal{D}_i$, reflecting variability in user behavior, local environments, or sensing conditions. That is, while traditional learning assumes all data points $x_i \sim \mathcal{D}$, FL encounters the more general scenario where:

\[
x_i \sim \mathcal{D}_i, \quad \text{for } x_i \in \text{client } i.
\]
As such, the global dataset is no longer exchangeable, and assumptions relying on this property break down.

Empirical and theoretical studies have shown that when clients violate exchangeability, such as by training on non-overlapping label distributions or domain-shifted data, the aggregated global model can suffer significant degradation in performance, particularly on the target distribution of interest \cite{lu2023federated}. This issue is exacerbated by the fact that clients’ local data distributions are not observable, either due to privacy constraints or system limitations. Consequently, we lack the information necessary to directly assess how well each client’s model update aligns with the target task.

\subsection{Malfunctioning Clients}
While violations of exchangeability across client updates arise from training on heterogeneous data distributions, leading to prediction errors as discussed above, malfunctioning clients represent a more extreme case of misalignment. In contrast to clients optimized for a specific data distribution, malfunctioning clients are generally misaligned with all data distributions.
In our work, we consider three distinct types of client malfunctions. Two of these represent untargeted model poisoning attacks, namely, the Additive-Noise Attack (ANA) and the Sign-Flipping Attack (SFA), which are widely studied attacks in the FL literature \cite{konstantin2024asmr,li2020learning,alebouyeh2024benchmarking}.

\textbf{In the ANA setting}, a client perturbs its model parameters by adding Gaussian noise before broadcasting, i.e., the transmitted update becomes:
$    \tilde{\theta} = \theta + \varepsilon, \quad \varepsilon \sim \mathcal{N}(0, \sigma^2I),
$ \\
where $\theta$ is the locally trained model and $\varepsilon$ is the noise vector.

\textbf{In the SFA setting}, the client multiplies the entire update by a negative constant, effectively reversing the optimization direction:
$\tilde{\theta} = -\alpha \cdot \theta, \quad \alpha > 0$.\\
In addition to adversarial behaviors, we simulate a malfunction indicative of technical failures, such as broken data pipelines or failed local training. In this scenario, the client submits a model update consisting of randomly initialized weights, entirely bypassing the optimization process.

\subsection{Error Decomposition}

As previously described, prediction errors may arise due to either malfunctions or the violation of exchangeability between the source and target distributions. In other words, if $x_i \sim \mathcal{D}_S$ during training and $x_j \sim \mathcal{D}_T$ during testing with $\mathcal{D}_S \ne \mathcal{D}_T$, then the exchangeability assumption is violated. The latter can be understood as natural shifts inherent to distributional differences, while malfunctions represent unnatural shifts introduced through adversarial or unexpected faulty behavior. Therefore, the overall prediction error of a corrupted model $h(\tilde{\theta})$ on a target distribution $T$ can be expressed as the combination of these two sources of error:

\begin{equation}
    \varepsilon(h(\tilde{\theta})) = \varepsilon_T(h(\theta)) + \varepsilon_{\mathcal{M}}(h(\tilde{\theta}))
\end{equation}
where \( \varepsilon_T(h(\theta)) \) is the error on the target domain as described in Eq. \ref{eq:domain_error}
and \( \varepsilon_{\mathcal{M}}(h({\tilde{\theta}})) \) is defined as 

\begin{equation}
    \varepsilon_{\mathcal{M}}(h({\tilde{\theta}})) = \mathbb{E}_{\mathbf{x} \sim D} \left[ |h(\theta;\mathbf{x}) - h(\tilde{\theta};\mathbf{x})| \right],
\end{equation}
the error of model $h$ due to corrupted parameters $\tilde{\theta}$.
This concept is visualized in Figure \ref{fig:errors}. To quantify the impact of this shift, the domain adaptation framework developed by Ben-David et al.\cite{ben2010theory} provides a theoretical upper bound on the expected target domain error of a model $h \in \mathcal{H}$,
trained on a source distribution $\mathcal{D}_S$, and evaluated on a target distribution $\mathcal{D}_T$. The bound is given as:
\begin{equation}
\varepsilon_T(h(\theta)) \leq \varepsilon_S(h(\theta)) + \frac{1}{2} d_{\mathcal{H} \Delta \mathcal{H}}(\mathcal{D}_S, \mathcal{D}_T) + \lambda,
\end{equation}
where $\varepsilon_T(h(\theta))$, $\varepsilon_S(h(\theta))$ are the expected errors of $h$ with parameters $\theta$ on the source distribution $\mathcal{D}_S$ and target distribution $\mathcal{D}_T$ as described in Eq. \ref{eq:domain_error},
$d_{\mathcal{H} \Delta \mathcal{H}}(\mathcal{D}_S, \mathcal{D}_T)$ is the classifier induced-divergence between the source and target distributions \cite{kifer2004detecting,ben2006analysis}, and 
$\lambda$ is the combined error of the ideal model on both domains, defined as:
    \begin{equation}
         \lambda = \min_{h(\theta) \in \mathcal{H}} \left[ \varepsilon_S(h(\theta)) + \varepsilon_T(h(\theta)) \right]
    \end{equation}
This bound demonstrates that the error on the target domain is influenced not only by the model’s performance on the source domain but also by the divergence between the source and target distributions.
Thus, the overall error of the corrupted model can be described as:

\begin{equation}
\varepsilon_T(h(\tilde{\theta})) \leq \underbrace{\varepsilon_S(h(\theta)) + \frac{1}{2} d_{\mathcal{H}\Delta\mathcal{H}}(D_S, D_T)+\lambda}_{\text{exchangeability}}+\underbrace{ \varepsilon_{\mathcal{M}}(h({\tilde{\theta}}))}_{\text{corruption}}
\label{eq:error}
\end{equation}
To assess the sensitivity to corruption of the models used in this work, we added an ablation study to the supplementary material that investigates how the prediction error changes with the level of corruption.

\subsection{Communication and Computation Tradeoffs}
In centralized FL, a server aggregates client updates without access to local data distributions, ideally selecting only those updates that align well with a target distribution. Since this target may differ across clients, the optimal aggregation set is inherently client-specific.
Switching from a star-shaped to a P2P topology allows clients to exchange updates directly, giving each client access to all updates and enabling them to estimate prediction error on their own data to form personalized aggregation sets. This decentralization increases communication costs, as updates must be shared with all clients, but offers improved robustness and personalization. To determine the aggregation set for each client individually, LIGHTYEAR analyzes the model’s prediction behavior. While this requires additional forward passes and Jacobian matrix computations and is therefore computationally more demanding, it enables more fine-grained control in assessing whether client updates are beneficial for aggregation.
In domains such as medical imaging, where client numbers are limited and reliability is critical, these tradeoffs are favorable. We added an extended analysis of the scalability and communication costs in the supplementary material.

\subsection{Approximation by Agreement}
What is available to client $i$ are: (1) its own locally trained model $h_i$, used as reference and representative of the local distribution, trained on data from $\mathcal{D}_i$, and (2) a local validation dataset $V_i = \{(x_k, y_k)\}_{k=1}^n$ drawn from a distribution $\mathcal{D}_{\text{val}} \approx \mathcal{D}_i$. 
Therefore, we need to reconsider the problem from another point of view. We assumed that a malfunctioning model will exhibit a higher prediction error on the target distribution compared to a model that was specifically optimized for it. This increased error manifests as a behavioral divergence between the two models when making predictions on the target domain. In particular, the corrupted model $h(\tilde{\theta})$ will tend to disagree with the optimal model $h(\theta^*)$, which was trained to minimize error on that specific distribution. This disagreement indicates the presence of a prediction error apart from the irreducible error inherent to the task, such that:

\begin{equation}
    \mathbb{E}_{\mathbf{x} \sim D_i} \left[ |h(\theta^*_i;\mathbf{x}) - h(\tilde{\theta}_j;\mathbf{x})| \right] > 0
\end{equation}

From this perspective, our goal is to identify a personalized aggregation set for each client, consisting only of client updates that exhibit strong agreement with the client’s local reference model. Let \( h_i := h(\theta_i) \) and \( h_j := h(\theta_j) \) denote the models of client \(i\) and client \(j\), respectively, where \( \theta_i \) and \( \theta_j \) are their respective parameter vectors. Since we evaluate models that have not yet converged during training, assessing performance alone may not provide meaningful insights. Instead, we aim to capture the tendencies of the models and therefore introduce our agreement score.

To identify neighboring clients whose predictive behavior is consistent with the local model, we compute an \emph{agreement score} based on the NTK induced by the final layer of the networks. We restrict the NTK computation to the final layer in order to significantly reduce the computational and memory complexity while still capturing the task-specific predictive behavior of the models. The NTK has already been shown to be an effective tool for model optimization in heterogeneous FL settings \cite{yue2022neural}. Intuitively, the NTK describes how the predictions of a neural network change around the data points when the model parameters are slightly modified. It quantifies how strongly and in which direction the model predictions change under small parameter perturbations and therefore captures the local behavior of the model rather than its raw parameter values. Consequently, models with similar NTK structure exhibit similar behavior in the function space, even if their parameter values differ.
For each model, we compute the Jacobian of the network outputs with respect to the final-layer parameters $\theta^{(L)}$,

\begin{equation}
J_{\theta}(x_i)
=
\frac{\partial h(\theta;x_i)}{\partial \theta^{(L)}}.
\end{equation}
The empirical final-layer NTK matrix is then defined as
\begin{equation}
[K_{\theta}(V_i)]_{ab}
=
\frac{1}{C}
\left\langle
J_{\theta}(x_a),
J_{\theta}(x_b)
\right\rangle,
\end{equation}
where $C$ denotes the output dimension of the model and $x \in V_i$. To remove scale dependencies and improve robustness, the kernel matrix is centered and Frobenius-normalized,
\begin{equation}
\tilde{K}_{\theta}(V_i)
=
\frac{H K_{\theta}(V_i) H}
{\|H K_{\theta}(V_i) H\|_F},
\qquad
H = I - \frac{1}{|V_i|}\mathbf{1}\mathbf{1}^\top.
\end{equation}
Given a local reference model $\theta_i$ and a neighboring model $\theta_j$, we define the NTK-based agreement score via normalized kernel alignment,
\begin{equation}
\mathcal{A}(\theta_i,\theta_j; V_i)
=
\left\langle
\tilde{K}_{\theta_i}(V_i),
\tilde{K}_{\theta_j}(V_i)
\right\rangle_F.
\end{equation}
Intuitively, this procedure favors models that exhibit similar local predictive sensitivities on the reference dataset and suppresses aggregation with models that rely on substantially different predictive mechanisms.
Since both kernel matrices are centered and Frobenius-normalized, the agreement score admits the following closed-form expression:
\begin{equation}
\mathcal{A}(\theta_i,\theta_j; V_i)
=
\frac{
\left\langle
H K_{\theta_i}(V_i) H,
H K_{\theta_j}(V_i) H
\right\rangle_F
}{
\|H K_{\theta_i}(V_i) H\|_F
\cdot
\|H K_{\theta_j}(V_i) H\|_F
}.
\end{equation}

\subsection{Update Selection}
Based on the agreement score defined in the previous section, we leverage a filtering mechanism that enables each client to selectively aggregate only those updates whose predictions are sufficiently aligned with its own reference model. We introduce a selection threshold $\tau \in \mathbb{R}$ and define the aggregation set $S_i$ for client $i$ as:

\begin{equation}
S_i = \{ \theta_j \in \mathcal{N}(i) \mid \mathcal{A}(\theta_i, \theta_j; V_i) \geq \tau \}.    
\end{equation}
This contains only those neighbour models that are considered sufficiently similar to the local model in terms of predictive behavior on the validation data. 

\subsection{Aggregation}
After selecting the aggregation set, the chosen updates are aggregated to obtain the new model. Since each client aggregates their own model during training, there is a risk of client drift. To enhance robustness, we introduce a regularization parameter into the aggregation process, as shown in Eq. \ref{eq:aggregation}. 

\begin{equation}
    \bar{\theta_i}^{(t+1)} = \bar{\theta_i}^{(t)} + \gamma^{t} \cdot \frac{1}{|\mathcal{S}_i|} \sum_{j \in \mathcal{S}_i} \left( \theta_j - \bar{\theta_i}^{(t)} \right)
    \label{eq:aggregation}
\end{equation}
This parameter is dependent on the training round and can be interpreted as a decay of change, reflecting our observation of a gradual performance decline after a certain number of training rounds. By incorporating this decay, the aggregation becomes more stable and resilient to fluctuations. We have conducted a corresponding ablation study to support our findings. Notably, setting the regulation parameter to 1 reduces the method to the standard FedAvg over the updates from the selected aggregation set.
\section{Datasets}
To evaluate the effectiveness of LIGHTYEAR, we conducted experiments on five diverse datasets. Each client holds an individual training, validation and test set, created through a random split of their local data. 

\textbf{FEMNIST} \cite{caldas2018leaf} contains $28 \times 28$ grayscale handwritten digits and characters across 62 classes. Data is partitioned by writer identity, giving each client a distinct distribution that naturally induces heterogeneity. We use 8 clients and train a two-layer CNN for this task.

\textbf{Camelyon17-WILDS} \cite{bandi2018detection} contains $98 \times 98$ tissue-slide patches for binary tumor classification, collected from five hospitals by different scanners. We assign one hospital per client, yielding five clients in total, each training a local DenseNet121 model \cite{huang2017densely}.

\textbf{Isic19} \cite{tschandl2018ham10000,codella2018skin,combalia2019bcn20000} contains dermoscopic images for multi-class skin cancer classification. Images are preprocessed to 
224 $\times$ 224 resolution and originate from six medical centers, yielding six clients. Each client trains an EfficientNet model \cite{tan2019efficientnet} following the common baseline setup \cite{NEURIPS2022_232eee8e}.

\textbf{Fetal Abdominal Structures (Ultrasound)} \cite{da2023fetal} is a binary segmentation task, with each patient assigned to a distinct set of patients across five clients. The images were resized to 64 $\times$ 64 and training was performed using TransUNet \cite{chen2024transunet}. 

\textbf{ChestXRay} \cite{johnson2019mimic} is a XRay segmentation task. For this paper, we used a subset of 200 samples distributed across five clients. As in the other segmentation task, the images are resized to 64 $\times$ 64 and TransUNet was used for training.
The detailed training setup and dataset composition are provided in the supplementary material.
\section{Experiments}\label{sec:experiments}
In the following experiments, we consider three different types of malfunctions: ANAs, SFAs and clients who submit random updates. For each experiment, we incrementally increase the number of malfunctioning clients, meaning that the number of malfunctioning clients grows per run. We compare LIGHTYEAR against FedAvg \cite{mcmahan2017communication} and eight baseline methods for robust aggregation: AFA \cite{munoz2019byzantine}, ASMR \cite{konstantin2024asmr}, CFL \cite{sattler2020byzantine}, Ditto \cite{li2021ditto}, Krum \cite{blanchard2017machine}, FedProx \cite{li2020federated}, BALANCE \cite{fang2024byzantine}, and SCCLIP \cite{yang2024byzantine}. While BALANCE and SCCLIP are also P2P methods, the others represent baselines from centralized FL. 
The performance of each method is evaluated based on the average accuracy for the classification tasks and the average Dice score for the segmentation tasks, evaluated on the client's test data. The experiments focus on three key scenarios: (1) evaluating the resilience of each method against the three types of malfunctions individually, (2) assessing the resilience based on topology as the number of malfunctioning clients increases, and (3) testing the resilience in a dynamic and highly unpredictable environment, where malfunctioning clients randomly select one of the three malfunctions each round. This dynamic scenario aims to simulate conditions that are more representative of real-world situations. For LIGHTYEAR, we set the hyperparameter $\gamma$ to 0.95 in all cases.   The decision threshold $\tau$ was set to 0.6 for all experiments. The choice of these hyperparameters is discussed in the ablation study.
\section{Results}
\label{sec:results}
The results demonstrate that LIGHTYEAR consistently outperforms all baseline approaches. We provide detailed tables containing the concrete numbers of our results in the supplementary material. As shown in Figure \ref{fig:box}, only LIGHTYEAR is able to deliver stable performance across all datasets under the evaluated training conditions.
\begin{figure}[]
    \centering
    \includegraphics[width=\linewidth]{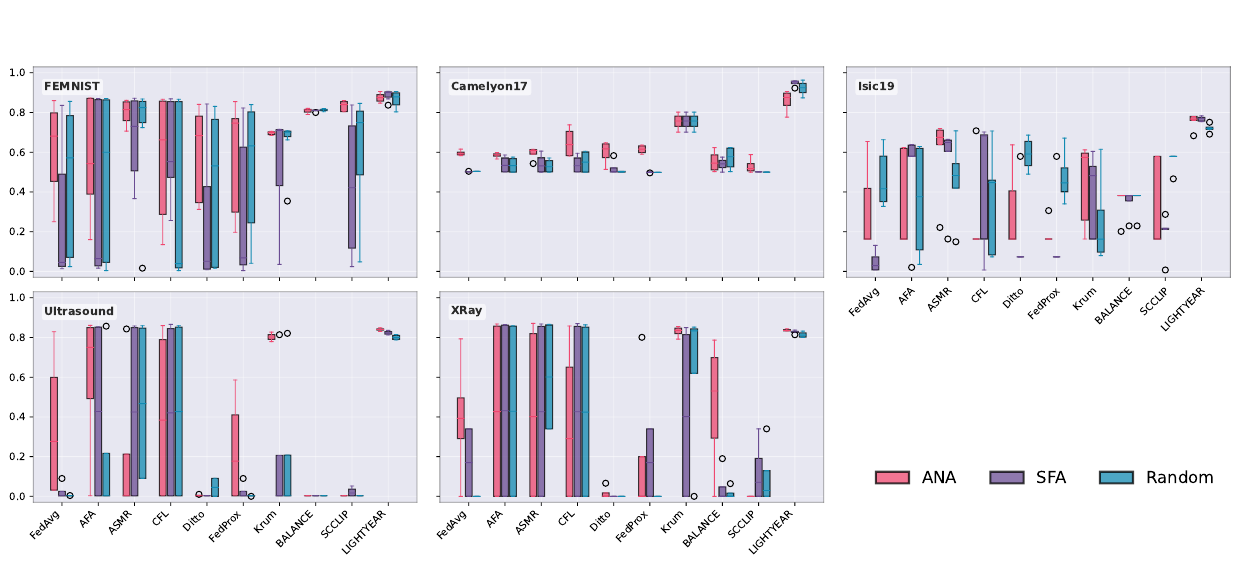}
    \caption{Average performance under three client malfunction types, aggregated over all cases with an increasing number of malfunctioning clients from 1 to $n\!-\!1$. Results report mean accuracy across clients and runs. Segmentation performance is reported as Dice score, while the classification performance is reported as accuracy.
}
    \label{fig:box}
\end{figure}
Especially for the segmentation tasks, it is evident that the models trained by the baseline methods diverge and deliver almost zero Dice scores, highlighting their inability to defend against any type of malfunctioning client. While certain cases, such as BALANCE on FEMNIST, occasionally achieve stable results, they remain exceptions rather than the norm.  Figure \ref{fig:topology} further illustrates the impact of an increasing number of clients, showing that all baseline methods encounter severe problems when the proportion of malfunctioning clients exceeds 50\%. Only in one case do the centralized FL methods deliver comparable performance.
\begin{figure}[]
    \centering
    \includegraphics[width=\linewidth]{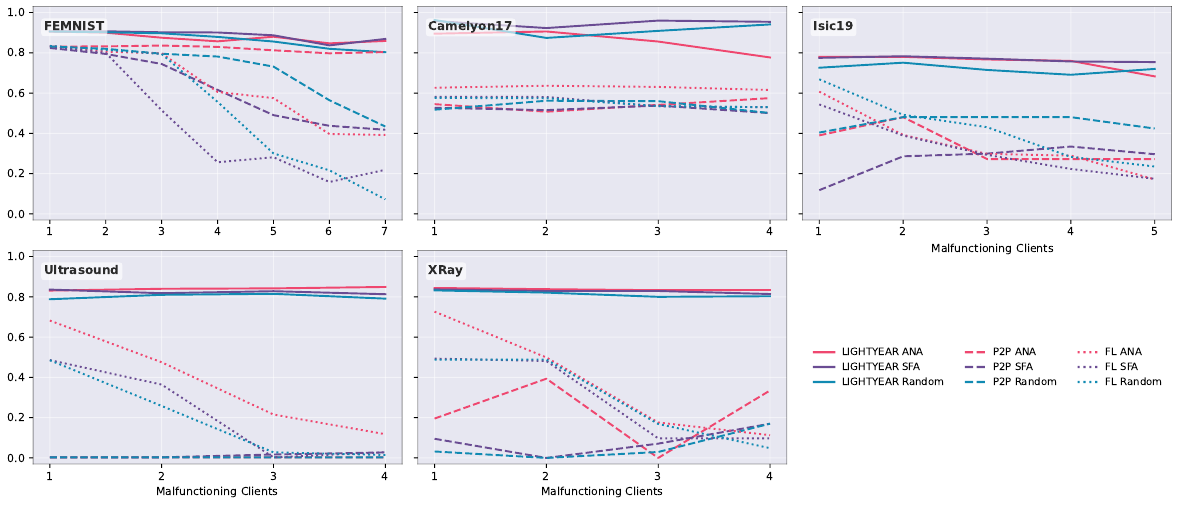}
    \caption{Performance under increasing numbers of malfunctioning clients for centralized FL, P2P FL, and LIGHTYEAR. Segmentation performance is reported as Dice score, while the classification performance is reported as accuracy.}
    \label{fig:topology}
\end{figure}
In contrast, LIGHTYEAR leverages access to the local data distribution to effectively identify and reject malfunctioning clients, maintaining stability even when they constitute the majority. Therefore, only LIGHTYEAR unlocks the full potential of the P2P topology. Finally, Figure \ref{fig:mix} highlights that LIGHTYEAR remains robust under dynamically changing malfunctioning clients, reflecting a more realistic, real-world scenario. 
\begin{figure}[]
    \centering
    \includegraphics[width=\linewidth]{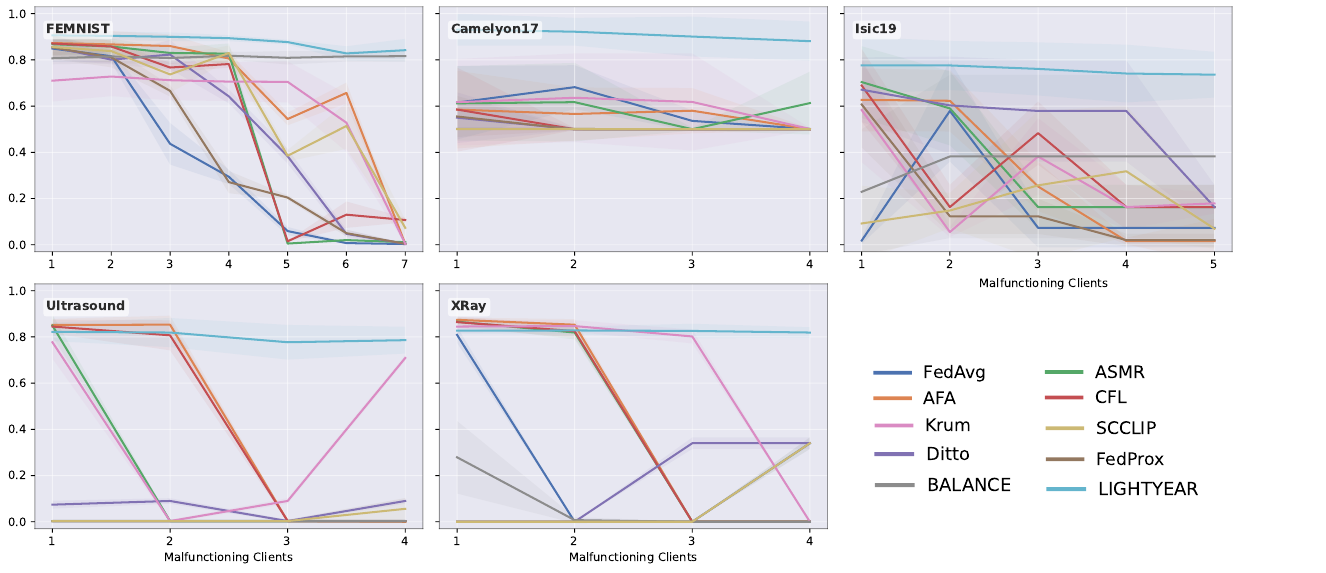}
    \caption{Illustrates the resilience of each method to dynamically changing client malfunctions. In this setting, each malfunctioning client randomly selects one of the three malfunction types in every round. The y-axis shows the average score (accuracy/Dice) across all clients, along with the standard deviation. Segmentation performance is reported as Dice score, while the classification performance is reported as accuracy.
}
    \label{fig:mix}
\end{figure}
Methods that perform relatively well on classification tasks often fail to maintain consistent performance on segmentation tasks. As a result, the other baseline approaches do not offer a reliable alternative for diverse medical imaging applications, highlighting the need for a method like LIGHTYEAR that remains robust across both task types.
For all experiments, we provide detailed information with exact numbers in the supplementary.
Beyond average performance, we observe that LIGHTYEAR exhibits substantially reduced variance across clients compared to both centralized and existing P2P baselines. This effect is particularly pronounced under high malfunction rates, where baseline methods either diverge or collapse to near-random performance. The agreement-based selection mechanism consistently filters out updates that induce unstable training dynamics, resulting in smoother convergence and more uniform client-level outcomes. These findings suggest that the primary advantage of LIGHTYEAR lies not only in higher mean performance, but also in improved reliability across heterogeneous clients.
\section{Ablation Study on $\gamma$ and $\tau$}

\begin{figure}[h]
    \centering
    \includegraphics[width=\linewidth]{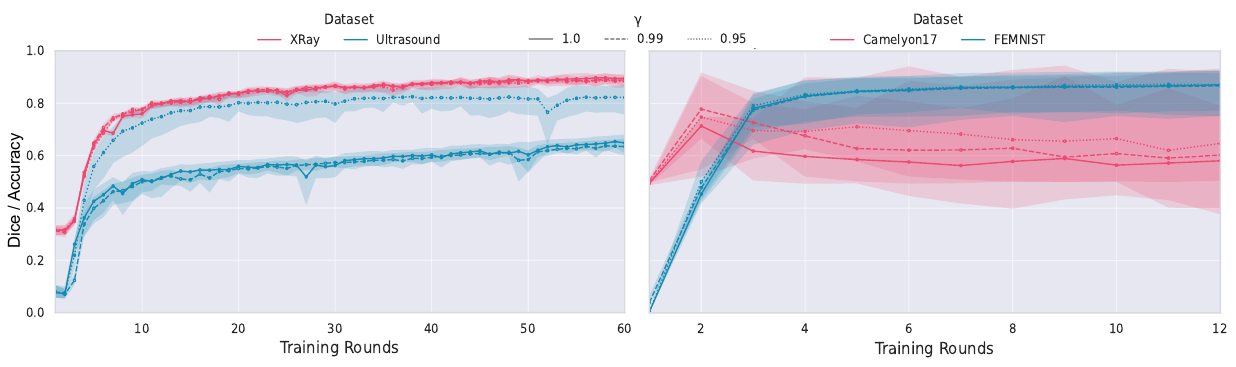}
    \caption{Effect of the regularization factor $\gamma$ measured by the average validation accuracy for classification and Dice for segmentation across all clients over training rounds.}
    \label{fig:ablation_gamma}
\end{figure}

\begin{figure}
    \centering
    \includegraphics[width=0.8\linewidth]{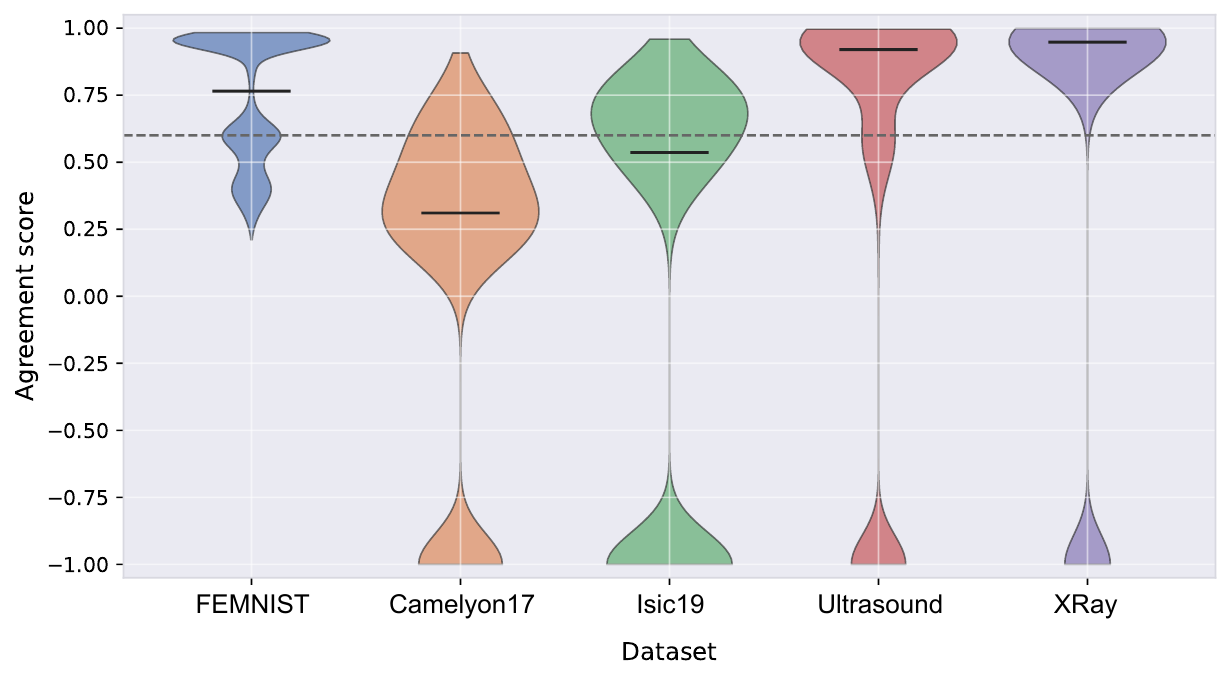}
    \caption{Distribution of the agreement score across all evaluated scenarios and clients, measured on the local validation datasets. The dashed line indicates the threshold $\tau=0.6$ used for update selection.}
    \label{fig:ablation_agreement}
\end{figure}

To study the effect of the regularization parameter $\gamma$, we isolated the regularized aggregation component and evaluated it independently of the personalized aggregation set selection. The corresponding results are shown in Figure \ref{fig:ablation_gamma}. Camelyon17 and FEMNIST represent the most distinct heterogeneity settings in our benchmark, with FEMNIST showing the lowest and Camelyon17 the highest degree of client heterogeneity. The results show that lower values of $\gamma$ stabilize training in highly heterogeneous settings, particularly on Camelyon17, while FEMNIST remains largely unaffected by changes in $\gamma$. A similar trend is observed for the segmentation datasets, where the ultrasound dataset benefits from a lower regularization factor of $\gamma = 0.95$, whereas the XRay dataset remains stable across all evaluated values. Based on these findings, we use $\gamma = 0.95$ for all experiments, as datasets either benefit from this choice or show little sensitivity to variations of $\gamma$.

Figure \ref{fig:ablation_agreement} shows the distribution of the agreement score across all evaluated scenarios. The dashed line indicates the threshold $\tau = 0.6$ used for update selection. A clear separation between malfunctioning and benign updates can be observed. The figure additionally reflects the degree of dataset heterogeneity, with Camelyon17 exhibiting the lowest agreement scores and therefore the strongest client heterogeneity. Since benign updates in highly heterogeneous settings can still yield lower agreement scores while remaining beneficial for the target domain, $\tau$ was chosen conservatively in the presented experiments.

\section{Conclusion}
In this work, we addressed the challenge of personalized update selection in FL under heterogeneous and unreliable client populations. A key observation of our work is that parameter-space similarity is often an insufficient proxy for model behavior, particularly in heterogeneous federated settings where small parameter changes can lead to substantially different predictions on local data distributions. This limits the reliability of conventional update selection strategies that operate solely in parameter space.
To overcome this limitation, we proposed LIGHTYEAR, a function-space-based framework for personalized and robust federated aggregation. At the core of LIGHTYEAR is a client-specific and NTK-based agreement score that estimates the predictive behavior of incoming updates on the local target domain. By evaluating semantic alignment in function space rather than parameter similarity, each client can identify the subset of updates that is most beneficial for its own data distribution. This enables a more expressive and behavior-aware update selection mechanism for FL.
The P2P topology plays a central role in enabling this perspective. Unlike centralized FL, where the server has access only to transmitted parameters, the decentralized setting allows each client to directly evaluate incoming updates on its own local validation data before aggregation. This provides access to the function space and enables personalized aggregation decisions based on model behavior rather than parameter heuristics. LIGHTYEAR further combines this update selection with a regularized aggregation strategy to stabilize training, improve robustness, and mitigate client drift.
Our experiments across multiple datasets and challenging heterogeneous and malfunctioning scenarios demonstrate that LIGHTYEAR consistently improves performance compared to both centralized FL methods and state-of-the-art P2P baselines. These findings highlight the value of function-space-based reasoning for federated optimization and suggest that future FL methods may benefit from moving beyond parameter-space metrics toward behavior-aware aggregation strategies. 

\bibliography{main}
\appendix

\section{Dataset Construction} \label{apx:datasets}
For each dataset we provide a csv-file containing the concrete splits. 
As mentioned in the main text, each client holds a unique training, validation, and test set. 

\begin{figure}
    \centering
    \includegraphics[width=\linewidth]{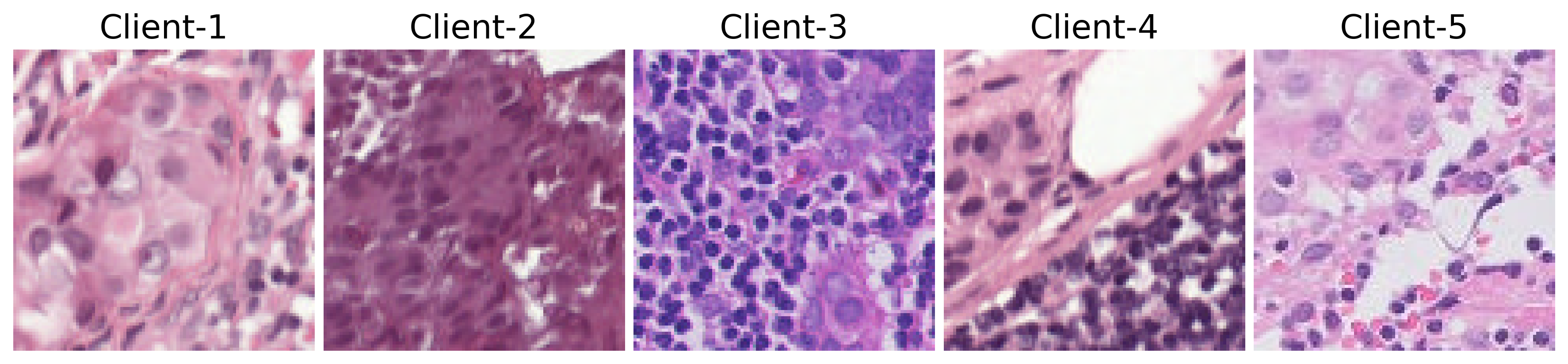}
    \caption{Sample images from each client, illustrating distinct differences in their data distributions. These variations highlight the heterogeneity across clients, which contributes to prediction errors on target domains as described in Eq 5 of the main paper, and underline the challenges of training in federated settings.}
    \label{fig:clients}
\end{figure}

We perform experiments on datasets of five different modalities, including three classification and two segmentation tasks.
We used the \textbf{Camelyon17-WILDS} \cite{bandi2018detection} dataset as a representative of histopathology
, which consists of 96$\times$96 H\&E-stained tissue slide patches of lymph nodes for a binary tumor classification task. The dataset contains images from patients across five hospitals, which naturally form three domains due to differences in image acquisition protocols namely, DHistech P250 (0.24 $\mu m$ pixel size), Philips IntelliSite Ultra Fast Scanner (0.25 $\mu m$), and Hamamatsu XR C12000 whole-slide scanner (0.23 $\mu m$). Each client holds the data of one medical center. The distribution shift across the clients is visualized in Figure \ref{fig:clients}. 

For the second dataset, \textbf{FEMNIST}, we used the dataset structure provided by the LEAF (https://github.com/TalwalkarLab/leaf) repository. In this setting, each client receives a distinct subset of data corresponding to a unique set of writers. Similar to the Camelyon17 setup, a random train/validation/test split is performed on each client’s data. Our preprocessing scripts for generating these subsets and splits are also included in the datasets directory of our codebase.

\begin{figure}
    \centering
    \includegraphics[width=\linewidth]{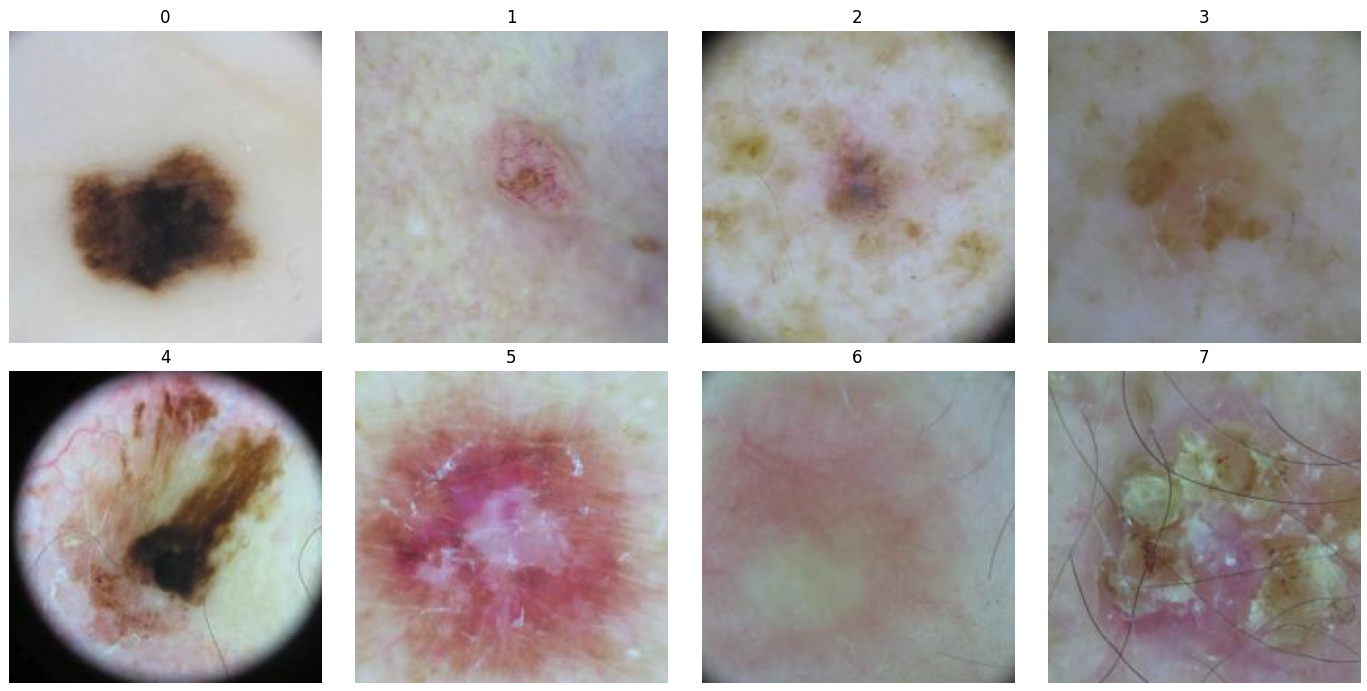}
    \caption{Sample images from each client, illustrating distinct differences in their data distributions. These variations highlight the heterogeneity across clients, which contributes to prediction errors on target domains as described in the main paper, and underline the challenges of training in federated settings.}
    \label{fig:classes}
\end{figure}

The \textbf{Isic19} dataset is a large-scale dermoscopic image collection designed for skin-lesion classification. It includes images labeled across eight diagnostic categories, ranging from benign lesions to malignant skin cancers. These classes form the basis of the multi-class classification task explored in this work. Representative examples of all eight classes are visualized in Figure \ref{fig:classes}. Since the images have different dimensions across the different centers, we performed a random $200 \times 200$ crop in out data loader. Further, we applied RandomScale, Rotate, RandomBrightnessContrast, AffineProjection, and Normalization augmentation to the images in the training loop.

\begin{figure}
    \centering
    \includegraphics[width=\linewidth]{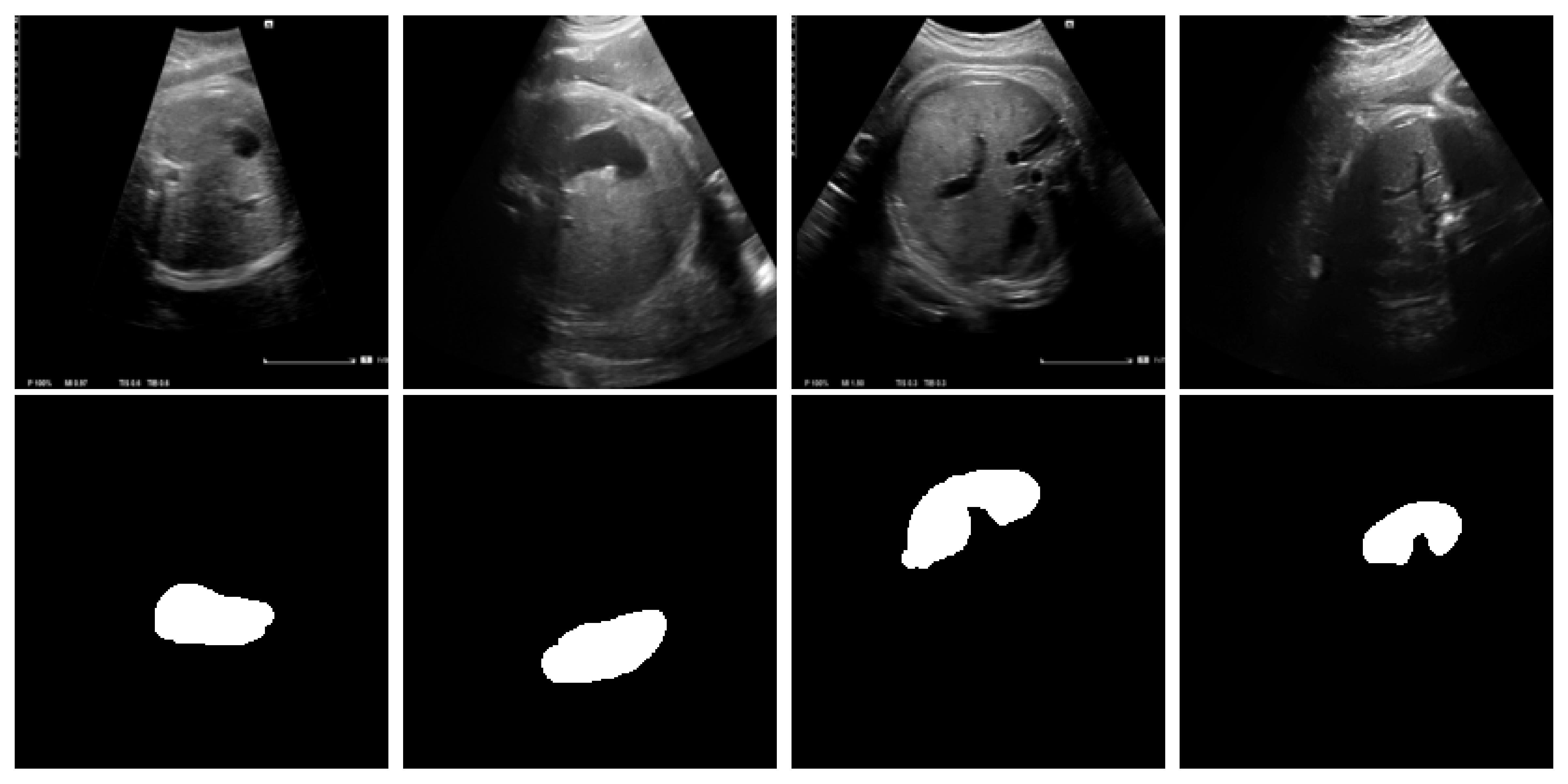}
    \caption{Sample images with the corresponding segmentation masks from the ultrasound dataset}
    \label{fig:us}
\end{figure}

We employ ultrasound scans from the Fetal Abdominal Structures Segmentation dataset \cite{da2023fetal}, which contains 1,588 fetal abdomen circumference images. These scans were acquired following a standardized protocol using Siemens Acuson, Voluson 730 (GE Healthcare Ultrasound), or Philips EPIQ Elite (Philips Healthcare Ultrasound) devices. In our experiments, we focus on segmenting the fetal liver. All images and corresponding segmentation masks are resized to $64 \times 64$. Each client receives a distinct subset of patients, forming its local dataset. Figure \ref{fig:us} shows some examples.

\begin{figure}
    \centering
    \includegraphics[width=\linewidth]{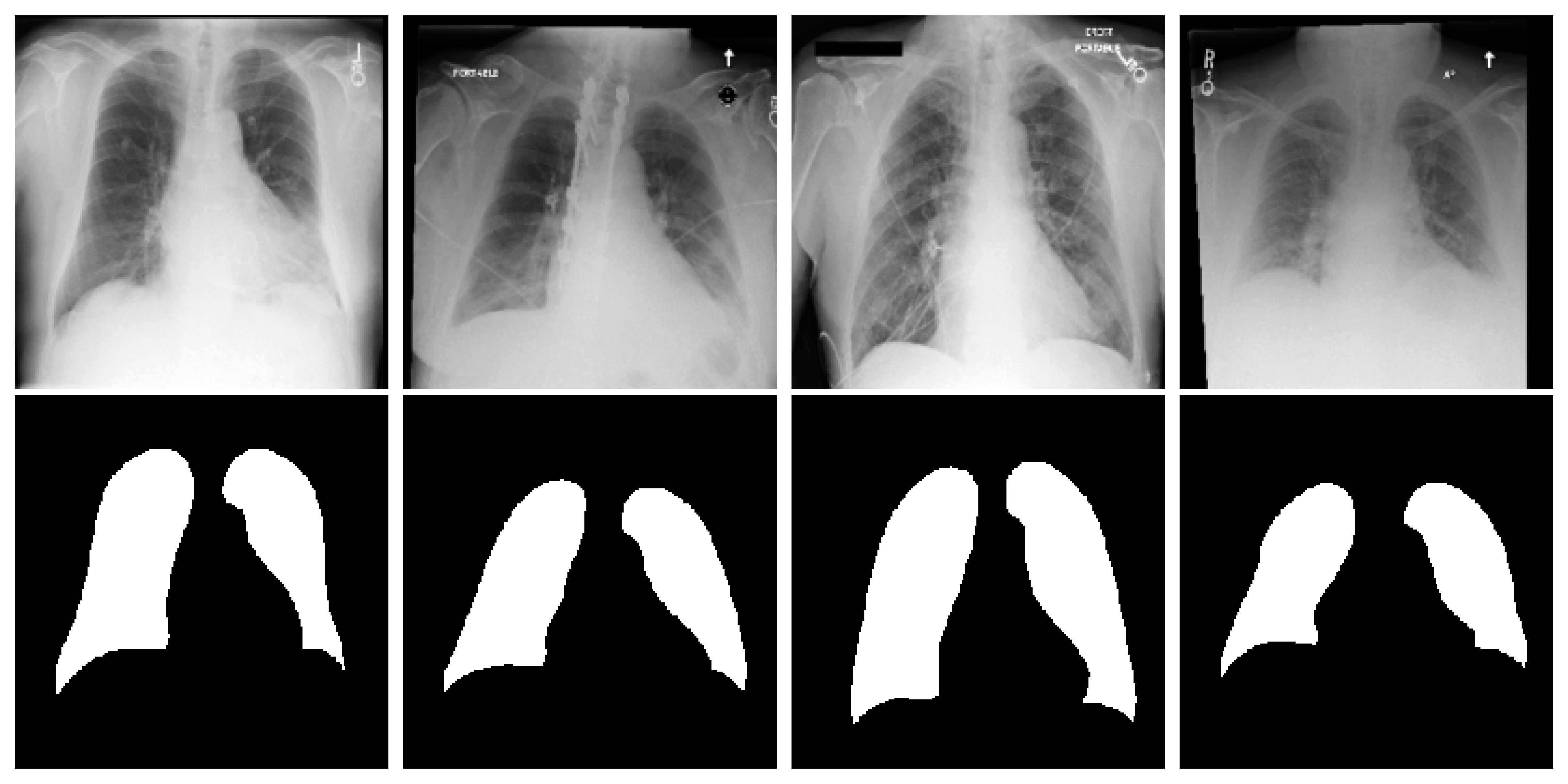}
    \caption{Sample images with the corresponding segmentation masks from the xray dataset}
    \label{fig:xray}
\end{figure}

Lastly, we use chest X-ray images from patients in tertiary care obtained from the MIMIC-XCR dataset \cite{johnson2019mimic}. Lung segmentation masks are sourced from the CheXmask database \cite{gaggion2024chexmask}. Because these masks were automatically generated, we retain only samples with a mean Dice Reverse Classification Accuracy of at least 70\%. For our experiments, we randomly select 200 images, resulting in 40 samples per client across five client datasets. All images and segmentation masks are resized to $64 \times 64$. Figure \ref{fig:xray} shows some images with their corresponding segmentation masks.

\section{Scalability Analysis}
As illustrated in Figure~\ref{fig:communication}, the communication complexity of centralized FL and P2P FL differs substantially with increasing federation size. In centralized FL, each client communicates exclusively with the central server by transmitting its local model update and receiving the aggregated global model. Consequently, the communication cost per client remains constant across rounds, so $\mathcal{O}(1)$, whereas the server-side communication cost increases linearly with the number of participating clients. Summed over the entire federation, this results in an overall communication complexity of $\mathcal{O}(n)$. In contrast, for the fully connected P2P topology considered in this work, P2P FL eliminates the central server and instead requires clients to exchange updates directly with one another. Under this topology, the communication cost per client increases linearly with the number of participating clients, meaning $\mathcal{O}(n)$, since each additional client introduces an additional communication partner. As a result, the total communication cost over the whole federation grows quadratically, so $\mathcal{O}(n^2)$. Figure~\ref{fig:communication} visualizes this difference for federations of varying size and highlights that, although P2P FL enables decentralized and client-specific decision making, this flexibility comes at the cost of substantially higher communication overhead compared to centralized FL.

\begin{figure}
\label{fig:communication}
    \centering
    \includegraphics[width=0.8\linewidth]{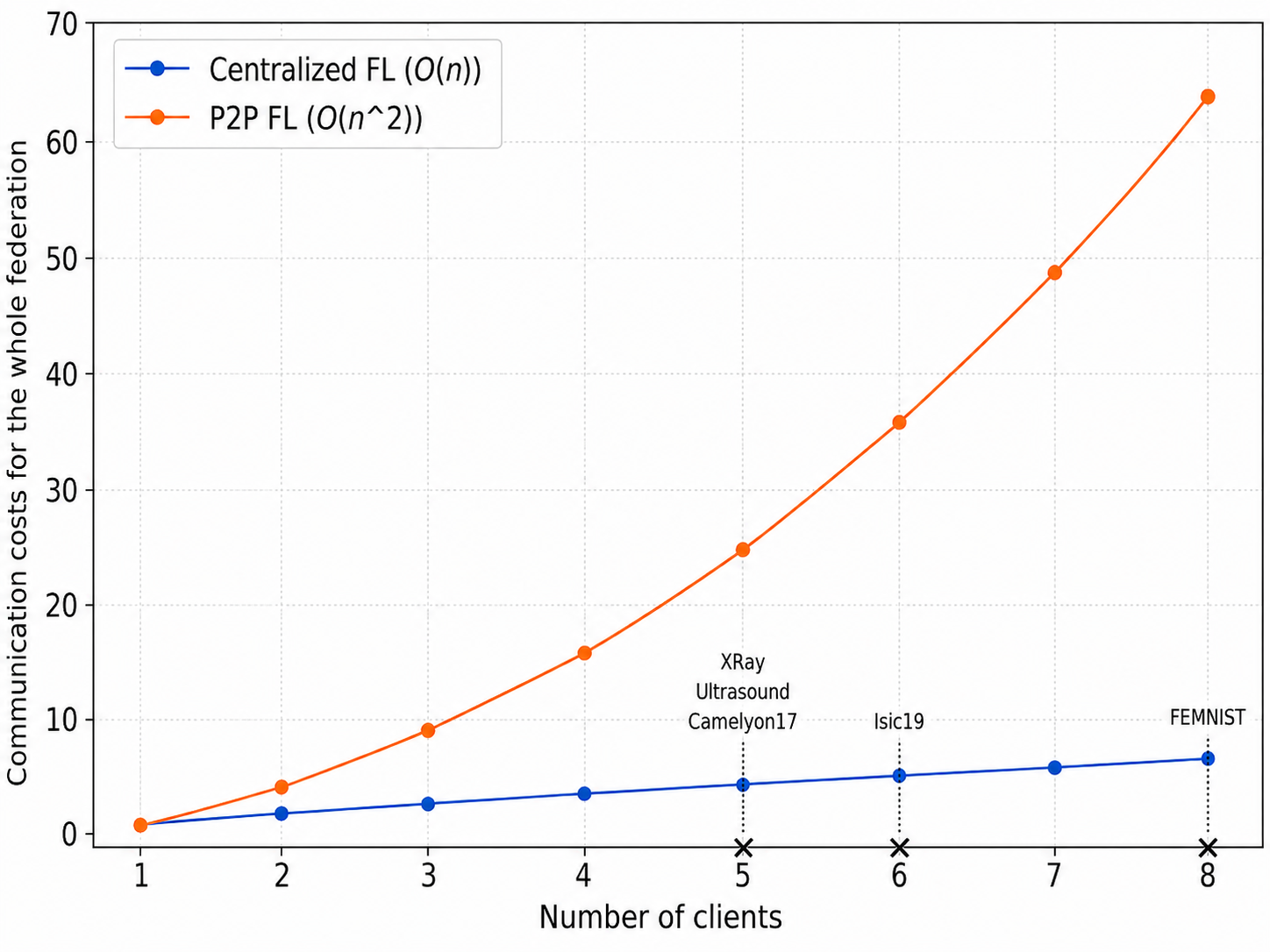}
    \caption{Comparison of communication costs between centralized and P2P topology as a function of the number of participating clients. While the total communication cost in centralized FL grows linearly with the federation size, $\mathcal{O}(n)$, the cost in P2P FL grows quadratically, $\mathcal{O}(n^2)$. The markers indicate the client counts used in our experiments for the respective datasets.}
    \label{fig:communication}
\end{figure}
\section{Experimental Results}
In this section, we provide a detailed breakdown of the results presented in the main paper. For each dataset and malfunction, a dedicated table is included. These tables report the average accuracy or dice score across all clients, along with the corresponding standard deviation. All values are expressed as percentages for consistency and clarity.

\begin{table*}[ht]
\centering
\caption{This table reports the results for ANAs on FEMNIST. Average accuracy (\%) $\pm$ std of each method under increasing number of malfunctioning clients.}
\vspace{0.5em}
\resizebox{\textwidth}{!}{%
\begin{tabular}{lccccccc}
\toprule
\multicolumn{8}{c}{\textbf{FEMNIST – ANA}} \\
\midrule
\multicolumn{8}{c}{\textbf{Number of Malfunctioning Clients}} \\
\cmidrule(lr){2-8}
\textbf{Method} & \textbf{1} & \textbf{2} & \textbf{3} & \textbf{4} & \textbf{5} & \textbf{6} & \textbf{7} \\
\midrule
FedAvg       & 86.0 $\pm$ 8.3 & 84.7 $\pm$ 7.3 & 75.0 $\pm$ 5.8 & 59.4 $\pm$ 5.5 & 68.1 $\pm$ 8.0 & 25.0 $\pm$ 1.2 & 31.2 $\pm$ 4.9 \\
AFA          & 87.2 $\pm$ 6.7 & 87.3 $\pm$ 6.5 & 87.0 $\pm$ 5.1 & 54.3 $\pm$ 3.9 & 52.7 $\pm$ 4.0 & 16.0 $\pm$ 3.9 & 25.0 $\pm$ 3.0 \\
ASMR         & 86.1 $\pm$ 4.8 & 85.5 $\pm$ 2.5 & 85.3 $\pm$ 2.1 & 81.5 $\pm$ 2.0 & 78.2 $\pm$ 5.3 & 73.6 $\pm$ 8.0 & 70.6 $\pm$ 8.5 \\
CFL          & 86.7 $\pm$ 5.8 & 86.0 $\pm$ 4.0 & 85.5 $\pm$ 2.6 & 13.5 $\pm$ 1.2 & 66.2 $\pm$ 3.6 & 36.2 $\pm$ 3.2 & 21.2 $\pm$ 1.6 \\
Ditto        & 84.1 $\pm$ 8.6 & 76.0 $\pm$ 4.7 & 80.3 $\pm$ 7.7 & 68.4 $\pm$ 1.9 & 32.5 $\pm$ 2.0 & 31.2 $\pm$ 2.3 & 36.7 $\pm$ 4.1 \\
FedProx      & 85.5 $\pm$ 7.2 & 76.2 $\pm$ 6.0 & 74.7 $\pm$ 2.7 & 77.6 $\pm$ 5.0 & 34.3 $\pm$ 4.4 & 25.3 $\pm$ 1.6 & 19.7 $\pm$ 4.1 \\
Krum         & 68.6 $\pm$ 9.4 & 68.5 $\pm$ 9.2 & 70.0 $\pm$ 9.1 & 69.2 $\pm$ 9.2 & 70.5 $\pm$ 8.7 & 70.5 $\pm$ 9.2 & 70.1 $\pm$ 9.1 \\
BALANCE      & 80.2 $\pm$ 2.8 & 81.3 $\pm$ 2.3 & 81.3 $\pm$ 1.6 & 81.9 $\pm$ 2.6 & 82.1 $\pm$ 2.2 & 79.0 $\pm$ 2.8 & 80.2 $\pm$ 3.4 \\
SCCLIP       & 86.0 $\pm$ 3.6 & 85.1 $\pm$ 3.7 & 85.9 $\pm$ 3.7 & 84.0 $\pm$ 3.4 & 80.4 $\pm$ 4.8 & 80.5 $\pm$ 3.1 & 80.5 $\pm$ 3.3 \\
LIGHTYEAR    & 90.7 $\pm$ 2.0 & 89.9 $\pm$ 2.4 & 87.5 $\pm$ 3.0 & 85.7 $\pm$ 6.7 & 88.0 $\pm$ 1.2 & 84.7 $\pm$ 1.5 & 81.4 $\pm$ 10.0 \\
\bottomrule
\end{tabular}}
\label{tab:femnist_ana}
\end{table*}

\begin{table*}[ht]
\centering
\caption{This table reports the results for SFAs on FEMNIST. Average accuracy (\%) $\pm$ std of each method under increasing number of malfunctioning clients.}
\vspace{0.5em}
\resizebox{\textwidth}{!}{%
\begin{tabular}{lccccccc}
\toprule
\multicolumn{8}{c}{\textbf{FEMNIST – SFA}} \\
\midrule
\multicolumn{8}{c}{\textbf{Number of Malfunctioning Clients}} \\
\cmidrule(lr){2-8}
\textbf{Method} & \textbf{1} & \textbf{2} & \textbf{3} & \textbf{4} & \textbf{5} & \textbf{6} & \textbf{7} \\
\midrule
FedAvg       & 83.6 $\pm$ 5.4 & 76.6 $\pm$ 2.4 & 21.4 $\pm$ 5.7 & 1.5 $\pm$ 0.3 & 1.3 $\pm$ 0.3 & 3.6 $\pm$ 0.4 & 4.5 $\pm$ 1.5 \\
AFA          & 86.9 $\pm$ 7.1 & 87.0 $\pm$ 6.1 & 86.0 $\pm$ 4.2 & 2.6 $\pm$ 0.3 & 6.4 $\pm$ 0.9 & 2.9 $\pm$ 0.3 & 1.6 $\pm$ 0.3 \\
ASMR         & 87.2 $\pm$ 4.6 & 86.8 $\pm$ 3.4 & 84.9 $\pm$ 1.7 & 73.0 $\pm$ 6.3 & 58.8 $\pm$ 4.6 & 42.5 $\pm$ 10.5 & 36.6 $\pm$ 2.9 \\
CFL          & 86.9 $\pm$ 4.8 & 85.9 $\pm$ 3.4 & 85.2 $\pm$ 3.2 & 25.6 $\pm$ 4.4 & 51.3 $\pm$ 12.5 & 55.3 $\pm$ 1.1 & 43.4 $\pm$ 10.0 \\
Ditto        & 84.3 $\pm$ 4.2 & 77.6 $\pm$ 1.9 & 7.7 $\pm$ 1.0 & 5.1 $\pm$ 0.5 & 1.4 $\pm$ 0.3 & 0.8 $\pm$ 0.2 & 0.9 $\pm$ 0.1 \\
FedProx      & 82.3 $\pm$ 4.0 & 74.3 $\pm$ 2.1 & 4.1 $\pm$ 1.6 & 0.4 $\pm$ 0.2 & 6.8 $\pm$ 1.0 & 2.4 $\pm$ 5.3 & 50.8 $\pm$ 10.0 \\
Krum         & 71.4 $\pm$ 8.8 & 71.5 $\pm$ 8.2 & 70.9 $\pm$ 9.0 & 71.4 $\pm$ 8.2 & 71.0 $\pm$ 8.7 & 3.5 $\pm$ 0.5 & 15.5 $\pm$ 3.2 \\
BALANCE      & 81.0 $\pm$ 2.5 & 80.0 $\pm$ 2.3 & 81.3 $\pm$ 2.4 & 81.1 $\pm$ 2.0 & 81.4 $\pm$ 2.3 & 80.9 $\pm$ 3.1 & 81.2 $\pm$ 3.0 \\
SCCLIP       & 83.8 $\pm$ 2.6 & 78.9 $\pm$ 3.3 & 67.6 $\pm$ 8.2 & 42.1 $\pm$ 16.1 & 16.7 $\pm$ 5.2 & 6.6 $\pm$ 5.0 & 2.4 $\pm$ 1.0 \\
LIGHTYEAR    & 90.6 $\pm$ 2.2 & 90.7 $\pm$ 2.0 & 90.2 $\pm$ 1.7 & 90.1 $\pm$ 1.5 & 88.7 $\pm$ 1.0 & 83.7 $\pm$ 2.7 & 86.9 $\pm$ 0.7 \\
\bottomrule
\end{tabular}}
\label{tab:femnist_ana}
\end{table*}

\begin{table*}[ht]
\centering
\caption{This table reports the results for the Random malfunction on FEMNIST. Average accuracy (\%) $\pm$ std of each method under increasing number of malfunctioning clients.}
\vspace{0.5em}
\resizebox{\textwidth}{!}{%
\begin{tabular}{lccccccc}
\toprule
\multicolumn{8}{c}{\textbf{FEMNIST – Random}} \\
\midrule
\multicolumn{8}{c}{\textbf{Number of Malfunctioning Clients}} \\
\cmidrule(lr){2-8}
\textbf{Method} & \textbf{1} & \textbf{2} & \textbf{3} & \textbf{4} & \textbf{5} & \textbf{6} & \textbf{7} \\
\midrule
FedAvg       & 85.6 $\pm$ 6.5 & 77.2 $\pm$ 3.0 & 79.6 $\pm$ 5.4 & 57.2 $\pm$ 3.8 & 10.0 $\pm$ 0.9 & 2.4 $\pm$ 0.4 & 4.2 $\pm$ 1.2 \\
AFA          & 87.0 $\pm$ 7.1 & 86.7 $\pm$ 6.3 & 86.0 $\pm$ 4.2 & 60.0 $\pm$ 3.8 & 5.0 $\pm$ 1.5 & 4.1 $\pm$ 1.6 & 0.4 $\pm$ 0.3 \\
ASMR         & 86.8 $\pm$ 3.9 & 86.0 $\pm$ 3.5 & 85.2 $\pm$ 1.8 & 82.5 $\pm$ 2.3 & 77.5 $\pm$ 5.4 & 72.4 $\pm$ 8.6 & 1.6 $\pm$ 0.7 \\
CFL          & 86.7 $\pm$ 4.9 & 85.8 $\pm$ 2.7 & 85.2 $\pm$ 2.2 & 3.9 $\pm$ 1.1 & 0.4 $\pm$ 0.2 & 0.6 $\pm$ 0.2 & 3.0 $\pm$ 0.1 \\
Ditto        & 83.1 $\pm$ 5.0 & 81.3 $\pm$ 5.7 & 71.8 $\pm$ 2.9 & 53.1 $\pm$ 2.6 & 2.3 $\pm$ 0.6 & 1.7 $\pm$ 0.7 & 1.6 $\pm$ 0.7 \\
FedProx      & 84.0 $\pm$ 5.2 & 82.7 $\pm$ 6.0 & 78.0 $\pm$ 1.8 & 63.2 $\pm$ 2.0 & 44.4 $\pm$ 13.6 & 4.1 $\pm$ 1.6 & 4.5 $\pm$ 1.6 \\
Krum         & 70.5 $\pm$ 8.8 & 70.8 $\pm$ 8.7 & 70.1 $\pm$ 8.9 & 69.5 $\pm$ 9.6 & 70.7 $\pm$ 9.0 & 66.2 $\pm$ 9.4 & 35.4 $\pm$ 13.3 \\
BALANCE      & 81.7 $\pm$ 2.8 & 80.4 $\pm$ 2.7 & 81.3 $\pm$ 2.3 & 81.4 $\pm$ 2.5 & 80.8 $\pm$ 2.5 & 81.2 $\pm$ 3.3 & 82.0 $\pm$ 1.6 \\
SCCLIP       & 84.6 $\pm$ 2.6 & 83.6 $\pm$ 3.5 & 77.8 $\pm$ 4.2 & 74.9 $\pm$ 3.3 & 65.5 $\pm$ 6.0 & 31.7 $\pm$ 13.9 & 4.8 $\pm$ 2.2 \\
LIGHTYEAR    & 90.6 $\pm$ 2.2 & 90.2 $\pm$ 2.1 & 89.9 $\pm$ 1.6 & 87.9 $\pm$ 1.3 & 85.5 $\pm$ 1.4 & 82.0 $\pm$ 0.1 & 80.3 $\pm$ 0.4 \\
\bottomrule
\end{tabular}}
\label{tab:femnist_ana}
\end{table*}

\begin{table*}[ht]
\centering
\caption{This table reports the results for the dynamically changing malfunction on FEMNIST. Average accuracy (\%) $\pm$ std of each method under increasing number of malfunctioning clients.}
\vspace{0.5em}
\resizebox{\textwidth}{!}{%
\begin{tabular}{lccccccc}
\toprule
\multicolumn{8}{c}{\textbf{FEMNIST – Dynamic Malfunctions}} \\
\midrule
\multicolumn{8}{c}{\textbf{Number of Malfunctioning Clients}} \\
\cmidrule(lr){2-8}
\textbf{Method} & \textbf{1} & \textbf{2} & \textbf{3} & \textbf{4} & \textbf{5} & \textbf{6} & \textbf{7} \\
\midrule
FedAvg       & 84.9 $\pm$ 5.8 & 81.6 $\pm$ 2.6 & 43.7 $\pm$ 9.0 & 29.4 $\pm$ 3.1 & 5.9 $\pm$ 1.8 & 0.7 $\pm$ 0.3 & 0.3 $\pm$ 0.2 \\
AFA          & 87.3 $\pm$ 7.2 & 86.7 $\pm$ 5.8 & 86.0 $\pm$ 5.0 & 80.6 $\pm$ 2.7 & 54.4 $\pm$ 2.7 & 65.7 $\pm$ 4.5 & 0.4 $\pm$ 0.2 \\
ASMR         & 86.8 $\pm$ 4.9 & 85.8 $\pm$ 3.4 & 83.0 $\pm$ 2.3 & 82.7 $\pm$ 4.4 & 0.5 $\pm$ 0.1 & 2.0 $\pm$ 0.6 & 1.1 $\pm$ 0.4 \\
CFL          & 87.1 $\pm$ 5.1 & 85.8 $\pm$ 2.4 & 76.7 $\pm$ 2.0 & 78.2 $\pm$ 1.6 & 1.5$\pm$ 0.2 & 13.0 $\pm$ 5.7 & 10.7 $\pm$ 1.0 \\
Ditto        & 85.6 $\pm$ 5.2 & 80.1 $\pm$ 2.0 & 82.3 $\pm$ 6.3 & 64.2 $\pm$ 3.0 & 38.3 $\pm$ 3.2 & 4.7 $\pm$ 1.8 & 0.4 $\pm$ 0.2 \\
FedProx      & 85.4 $\pm$ 5.0 & 81.1 $\pm$ 4.3 & 66.6 $\pm$ 2.5 & 27.1 $\pm$ 4.8 & 20.4 $\pm$ 1.0 & 5.0 $\pm$ 1.5 & 0.4 $\pm$ 0.2 \\
Krum         & 71.0 $\pm$ 9.0 & 72.8 $\pm$ 8.5 & 71.2 $\pm$ 8.7 & 70.6 $\pm$ 8.4 & 70.5 $\pm$ 8.9 & 52.8 $\pm$ 12.5 & 0.4 $\pm$ 0.2 \\
BALANCE      & 80.7 $\pm$ 2.6 & 81.5 $\pm$ 3.0 & 80.8 $\pm$ 2.6 & 81.8 $\pm$ 1.6 & 80.9 $\pm$ 2.9 & 81.5 $\pm$ 2.2 & 81.6 $\pm$ 2.3 \\
SCCLIP       & 85.6 $\pm$ 4.8 & 83.8 $\pm$ 4.8 & 73.7 $\pm$ 7.9 & 83.0 $\pm$ 3.2 & 38.5 $\pm$ 2.5 & 51.4 $\pm$ 9.9 & 7.3 $\pm$ 3.6 \\
LIGHTYEAR    & 90.6 $\pm$ 2.2 & 90.5 $\pm$ 1.8 & 89.9 $\pm$ 1.9 & 89.4 $\pm$ 1.3 & 87.7 $\pm$ 1.0 & 82.8 $\pm$ 3.1 & 84.2 $\pm$ 4.9 \\
\bottomrule
\end{tabular}}
\label{tab:femnist_ana}
\end{table*}

\begin{table*}[ht]
\centering
\caption{This table reports the results for ANAs on Camelyon17. Average accuracy (\%) $\pm$ std of each method under increasing number of malfunctioning clients.}
\vspace{0.5em}
\begin{tabular}{lcccc}
\toprule
\multicolumn{5}{c}{\textbf{Camelyon17 – ANA}} \\
\midrule
\multicolumn{5}{c}{\textbf{Number of Malfunctioning Clients}} \\
\cmidrule(lr){2-5}
\textbf{Method} & \textbf{1} & \textbf{2} & \textbf{3} & \textbf{4} \\
\midrule
FedAvg       & 58.8 $\pm$ 14.2 & 61.6 $\pm$ 18.2 & 59.5 $\pm$ 17.6 & 58.3 $\pm$ 18.5 \\
AFA          & 59.8 $\pm$ 19.0 & 58.5 $\pm$ 17.5 & 59.0 $\pm$ 15.1 & 56.6 $\pm$ 19.1 \\
ASMR         & 54.3 $\pm$ 10.3 & 61.5 $\pm$ 17.3 & 61.3 $\pm$ 14.9 & 60.6 $\pm$ 20.0 \\
CFL          & 58.3 $\pm$ 16.8 & 58.1 $\pm$ 18.2 & 69.4 $\pm$ 16.9 & 73.8 $\pm$ 18.8 \\
Ditto        & 64.1 $\pm$ 15.4 & 64.7 $\pm$ 16.1 & 59.3 $\pm$ 17.0 & 51.4 $\pm$ 0.5 \\
Krum         & 80.2 $\pm$ 0.4 & 77.5 $\pm$ 0.5 & 74.0 $\pm$ 0.6 & 70.1 $\pm$ 0.7 \\
FedProx      & 63.1 $\pm$ 16.3 & 63.6 $\pm$ 17.0 & 59.0 $\pm$ 13.9 & 60.0 $\pm$ 19.5 \\
BALANCE      & 50.2 $\pm$ 0.4 & 51.5 $\pm$ 2.0 & 57.5 $\pm$ 15.9 & 62.3 $\pm$ 15.9 \\
SCCLIP       & 58.8 $\pm$ 11.3 & 50.0 $\pm$ 0.4 & 50.8 $\pm$ 1.6 & 52.7 $\pm$ 5.5 \\
LIGHTYEAR    & 89.5 $\pm$ 8.8 & 90.6 $\pm$ 8.4 & 85.9 $\pm$ 15.1 & 77.7 $\pm$ 11.7 \\
\bottomrule
\end{tabular}
\label{tab:femnist_ana_short}
\end{table*}

\begin{table*}[ht]
\centering
\caption{This table reports the results for SFAs on Camelyon17. Average accuracy (\%) $\pm$ std of each method under increasing number of malfunctioning clients.}
\vspace{0.5em}
\begin{tabular}{lcccc}
\toprule
\multicolumn{5}{c}{\textbf{Camelyon17 – SFA}} \\
\midrule
\multicolumn{5}{c}{\textbf{Number of Malfunctioning Clients}} \\
\cmidrule(lr){2-5}
\textbf{Method} & \textbf{1} & \textbf{2} & \textbf{3} & \textbf{4} \\
\midrule
FedAvg       & 50.1 $\pm$ 0.4 & 50.1 $\pm$ 0.4 & 50.1 $\pm$ 0.4 & 50.4 $\pm$ 0.4 \\
AFA          & 56.6 $\pm$ 19.1 & 58.6 $\pm$ 17.7 & 50.1 $\pm$ 0.4 & 50.1 $\pm$ 0.4 \\
ASMR         & 56.1 $\pm$ 17.9 & 60.6 $\pm$ 17.8 & 50.0 $\pm$ 0.4 & 50.1 $\pm$ 0.4 \\
CFL          & 56.3 $\pm$ 14.4 & 59.5 $\pm$ 17.4 & 50.1 $\pm$ 0.4 & 50.1 $\pm$ 0.4 \\
Ditto        & 58.3 $\pm$ 10.8 & 50.1 $\pm$ 0.4 & 50.1 $\pm$ 0.4 & 50.1 $\pm$ 0.4 \\
FedProx      & 49.6 $\pm$ 3.3 & 50.1 $\pm$ 0.4 & 49.9 $\pm$ 0.4 & 49.9 $\pm$ 0.4 \\
Krum         & 80.2 $\pm$ 0.4 & 77.5 $\pm$ 0.5 & 74.0 $\pm$ 0.6 & 70.1 $\pm$ 0.7 \\
BALANCE      & 55.3 $\pm$ 7.4 & 53.0 $\pm$ 6.2 & 57.5 $\pm$ 14.3 & 50.0 $\pm$ 0.4 \\
SCCLIP       & 50.1 $\pm$ 0.4 & 50.1 $\pm$ 0.4 & 50.1 $\pm$ 0.4 & 50.1 $\pm$ 0.4 \\
LIGHTYEAR    & 95.8 $\pm$ 1.9 & 92.3 $\pm$ 6.2 & 96.0 $\pm$ 2.9 & 95.4 $\pm$ 2.8 \\
\bottomrule
\end{tabular}
\label{tab:femnist_ana_short}
\end{table*}

\begin{table*}[ht]
\centering
\caption{This table reports the results for Random malfunction on Camelyon17. Average accuracy (\%) $\pm$ std of each method under increasing number of malfunctioning clients.}
\vspace{0.5em}
\begin{tabular}{lcccc}
\toprule
\multicolumn{5}{c}{\textbf{Camelyon17 – Random}} \\
\midrule
\multicolumn{5}{c}{\textbf{Number of Malfunctioning Clients}} \\
\cmidrule(lr){2-5}
\textbf{Method} & \textbf{1} & \textbf{2} & \textbf{3} & \textbf{4} \\
\midrule
FedAvg       & 50.4 $\pm$ 0.4 & 50.4 $\pm$ 0.4 & 50.4 $\pm$ 0.4 & 50.4 $\pm$ 0.4 \\
AFA          & 57.5 $\pm$ 18.3 & 56.5 $\pm$ 20.0 & 50.0 $\pm$ 0.4 & 50.0 $\pm$ 0.4 \\
ASMR         & 55.3 $\pm$ 11.8 & 57.1 $\pm$ 17.5 & 50.0 $\pm$ 0.4 & 50.1 $\pm$ 0.4 \\
CFL          & 60.0 $\pm$ 16.6 & 60.5 $\pm$ 14.6 & 50.0 $\pm$ 0.4 & 50.0 $\pm$ 0.4 \\
Ditto        & 50.0 $\pm$ 0.4 & 50.0 $\pm$ 0.4 & 50.4 $\pm$ 0.4 & 50.4 $\pm$ 0.4 \\
FedProx      & 49.9 $\pm$ 0.4 & 49.9 $\pm$ 0.4 & 49.9 $\pm$ 0.4 & 49.9 $\pm$ 0.4 \\
Krum         & 80.2 $\pm$ 0.4 & 77.5 $\pm$ 0.5 & 74.0 $\pm$ 0.6 & 70.1 $\pm$ 0.7 \\
BALANCE      & 53.5 $\pm$ 6.6 & 62.4 $\pm$ 16.8 & 62.0 $\pm$ 14.9 & 50.2 $\pm$ 0.4 \\
SCCLIP       & 50.0 $\pm$ 0.4 & 50.0 $\pm$ 0.4 & 50.0 $\pm$ 0.4 & 50.0 $\pm$ 0.4 \\
LIGHTYEAR    & 96.3 $\pm$ 2.2 & 87.4 $\pm$ 6.9 & 90.9 $\pm$ 7.0 & 94.1 $\pm$ 4.3 \\
\bottomrule
\end{tabular}
\label{tab:femnist_ana_short}
\end{table*}

\begin{table*}[ht]
\centering
\caption{This table reports the results for the dynamically changing malfunction on Camelyon17. Average accuracy (\%) $\pm$ std of each method under an increasing number of malfunctioning clients.}
\vspace{0.5em}
\begin{tabular}{lcccc}
\toprule
\multicolumn{5}{c}{\textbf{Camelyon17 – Dynamic Malfunctions}} \\
\midrule
\multicolumn{5}{c}{\textbf{Number of Malfunctioning Clients}} \\
\cmidrule(lr){2-5}
\textbf{Method} & \textbf{1} & \textbf{2} & \textbf{3} & \textbf{4} \\
\midrule
FedAvg       & 61.5 $\pm$ 15.9 & 68.2 $\pm$ 9.5 & 53.6 $\pm$ 4.5 & 50.1 $\pm$ 0.4 \\
AFA          & 58.5 $\pm$ 16.8 & 56.6 $\pm$ 11.5 & 58.0 $\pm$ 9.8 & 50.0 $\pm$ 0.4 \\
ASMR         & 61.2 $\pm$ 16.1 & 61.7 $\pm$ 17.0 & 50.0 $\pm$ 0.6 & 61.3 $\pm$ 13.6 \\
CFL          & 58.5 $\pm$ 18.2 & 50.0 $\pm$ 0.4 & 50.0 $\pm$ 0.4 & 50.0 $\pm$ 0.4 \\
Ditto        & 55.0 $\pm$ 11.0 & 50.0 $\pm$ 0.4 & 50.0 $\pm$ 0.4 & 50.0 $\pm$ 0.4 \\
FedProx      & 55.5 $\pm$ 8.9 & 49.9 $\pm$ 0.4 & 49.9 $\pm$ 0.4 & 49.9 $\pm$ 0.4 \\
Krum         & 61.7 $\pm$ 18.7 & 63.6 $\pm$ 19.2 & 61.8 $\pm$ 21.0 & 50.1 $\pm$ 0.4 \\
BALANCE      & 50.1 $\pm$ 0.4 & 50.1 $\pm$ 0.4 & 50.0 $\pm$ 0.4 & 50.0 $\pm$ 0.4 \\
SCCLIP       & 50.0 $\pm$ 0.4 & 50.1 $\pm$ 0.4 & 50.0 $\pm$ 0.4 & 50.0 $\pm$ 0.4 \\
LIGHTYEAR    & 93.1 $\pm$ 6.9 & 92.2 $\pm$ 6.0 & 90.9 $\pm$ 8.6 & 88.1 $\pm$ 8.5 \\
\bottomrule
\end{tabular}
\label{tab:femnist_ana_short}
\end{table*}

\begin{table*}[ht]
\centering
\caption{This table reports the results for the dynamically changing malfunction on Isic19. Average accuracy (\%) $\pm$ std of each method under an increasing number of malfunctioning clients.}
\vspace{0.5em}
\resizebox{\textwidth}{!}{%
\begin{tabular}{lccccc}
\toprule
\multicolumn{6}{c}{\textbf{Isic19 – ANA}} \\
\midrule
\multicolumn{6}{c}{\textbf{Number of Malfunctioning Clients}} \\
\cmidrule(lr){2-6}
\textbf{Method} & \textbf{1} & \textbf{2} & \textbf{3} & \textbf{4} & \textbf{5}\\
\midrule
FedAvg       & 65.4 $\pm$ 15.8 & 41.8 $\pm$ 12.2 & 16.3 $\pm$ 9.6 & 16.3 $\pm$ 9.6 & 16.3 $\pm$ 16.3 \\
AFA          & 62.3 $\pm$ 12.1 & 61.9 $\pm$ 12.3 & 16.3 $\pm$ 9.6 & 16.3 $\pm$ 9.6  & 16.3 $\pm$ 16.3\\
ASMR         & 71.1 $\pm$ 15.7 & 71.9 $\pm$ 15.0 & 67.5 $\pm$ 12.0 & 63.8 $\pm$ 16.3 & 22.1 $\pm$ 13.2\\
CFL          & 70.8 $\pm$ 16.5 & 16.3 $\pm$ 9.6 & 16.3 $\pm$ 9.6 & 16.3 $\pm$ 9.6 & 16.3 $\pm$ 9.6\\
Ditto        & 63.7 $\pm$ 18.5 & 40.6 $\pm$ 9.2 & 16.3 $\pm$ 9.6 & 16.3 $\pm$ 9.6 & 16.3 $\pm$ 9.6\\
FedProx      & 30.6 $\pm$ 10.0 & 16.3 $\pm$ 9.6 & 16.3 $\pm$ 9.6 & 16.3 $\pm$ 9.6 & 16.3 $\pm$ 9.6\\
Krum         & 61.2 $\pm$ 20.3 & 25.8 $\pm$ 9.4 & 59.6 $\pm$ 22.2 & 57.4 $\pm$ 22.4 & 16.3 $\pm$ 9.6\\
BALANCE      & 20.1 $\pm$ 27.8 & 38.2 $\pm$ 35.3 & 38.2 $\pm$ 35.3 & 38.2 $\pm$ 35.3 & 38.2 $\pm$ 35.3\\
SCCLIP       & 57.9 $\pm$ 21.9 & 57.9 $\pm$ 21.9 & 16.3 $\pm$ 9.6 & 16.3 $\pm$ 9.6 & 16.3 $\pm$ 9.6\\
LIGHTYEAR    & 78.0 $\pm$ 9.3 & 78.0 $\pm$ 9.6 & 76.7 $\pm$ 11.1 & 76.6 $\pm$ 11.8 & 68.3 $\pm$ 9.2\\
\bottomrule
\end{tabular}}
\label{tab:femnist_ana_short}
\end{table*}

\begin{table*}[ht]
\centering
\caption{This table reports the results for the dynamically changing malfunction on Isic19. Average accuracy (\%) $\pm$ std of each method under an increasing number of malfunctioning clients.}
\vspace{0.5em}
\resizebox{\textwidth}{!}{%
\begin{tabular}{lccccc}
\toprule
\multicolumn{6}{c}{\textbf{Isic19 – SFA}} \\
\midrule
\multicolumn{6}{c}{\textbf{Number of Malfunctioning Clients}}\\
\cmidrule(lr){2-6}
\textbf{Method} & \textbf{1} & \textbf{2} & \textbf{3} & \textbf{4} & \textbf{5}\\
\midrule
FedAvg       & 13.1 $\pm$ 6.9 & 2.9 $\pm$ 2.7 & 7.3 $\pm$ 8.5 & 0.7 $\pm$ 0.9 & 0.7 $\pm$ 0.4 \\
AFA          & 63.6 $\pm$ 15.7 & 63.8 $\pm$ 12.6 & 2.0 $\pm$ 2.7 & 57.9 $\pm$ 21.9  & 57.9 $\pm$ 21.9\\
ASMR         & 66.5 $\pm$ 15.0 & 60.4 $\pm$ 17.7 & 64.9 $\pm$ 16.7 & 66.1 $\pm$ 17.4 & 16.3 $\pm$ 9.6\\
CFL          & 68.7 $\pm$ 16.4 & 70.2 $\pm$ 14.4 & 68.2 $\pm$ 15.2 & 0.7 $\pm$ 0.9 & 16.3 $\pm$ 9.6\\
Ditto        & 57.9 $\pm$ 21.9 & 7.3 $\pm$ 8.5 & 7.3 $\pm$ 8.5 & 7.3 $\pm$ 8.5 & 7.3 $\pm$ 8.5\\
FedProx      & 57.9 $\pm$ 21.9 & 7.3 $\pm$ 8.5 & 7.3 $\pm$ 8.5 & 7.3 $\pm$ 8.5 & 7.3 $\pm$ 8.5\\
Krum         & 52.9 $\pm$ 12.0 & 60.4 $\pm$ 19.5 & 48.2 $\pm$ 15.9 & 16.3 $\pm$ 9.6 & 16.3 $\pm$ 9.6\\
BALANCE      & 22.9 $\pm$ 27.1 & 35.5 $\pm$ 37.5 & 38.2 $\pm$ 35.3 & 38.2 $\pm$ 35.3 & 38.2 $\pm$ 35.3\\
SCCLIP       & 0.7 $\pm$ 0.9 & 21.7 $\pm$ 30.8 & 21.7 $\pm$ 33.3 & 28.7 $\pm$ 32.8 & 21.1 $\pm$ 33.3\\
LIGHTYEAR    & 77.5 $\pm$ 11.5 & 78.3 $\pm$ 10.0 & 77.1 $\pm$ 12.3 & 75.7 $\pm$ 12.3 & 75.4 $\pm$ 11.7\\
\bottomrule
\end{tabular}}
\label{tab:femnist_ana_short}
\end{table*}

\begin{table*}[ht]
\centering
\caption{This table reports the results for the dynamically changing malfunction on Isic19. Average accuracy (\%) $\pm$ std of each method under an increasing number of malfunctioning clients.}
\vspace{0.5em}
\resizebox{\textwidth}{!}{%
\begin{tabular}{lccccc}
\toprule
\multicolumn{6}{c}{\textbf{Isic19 – Random}} \\
\midrule
\multicolumn{6}{c}{\textbf{Number of Malfunctioning Clients}} \\
\cmidrule(lr){2-6}
\textbf{Method} & \textbf{1} & \textbf{2} & \textbf{3} & \textbf{4} & \textbf{5}\\
\midrule
FedAvg       & 66.3 $\pm$ 14.5 & 58.0 $\pm$ 13.5 & 41.6 $\pm$ 8.9 & 32.7 $\pm$ 15.2 & 35.1 $\pm$ 16.0 \\
AFA          & 62.9 $\pm$ 14.1 & 62.0 $\pm$ 14.5 & 37.6 $\pm$ 10.5 & 10.8 $\pm$ 1.4  & 3.5 $\pm$ 2.6\\
ASMR         & 70.8 $\pm$ 15.3 & 54.3 $\pm$ 19.5 & 41.9 $\pm$ 12.7 & 48.3 $\pm$ 19.6 & 14.9 $\pm$ 2.6\\
CFL          & 70.7 $\pm$ 14.3 & 44.7 $\pm$ 14.3 & 45.9 $\pm$ 15.6 & 7.3 $\pm$ 2.6 & 8.3 $\pm$ 3.9\\
Ditto        & 68.6 $\pm$ 17.8 & 65.5 $\pm$ 16.1 & 59.1 $\pm$ 17.1 & 49.0 $\pm$ 12.3 & 53.2 $\pm$ 13.3\\
FedProx      & 67.1 $\pm$ 13.2 & 52.1 $\pm$ 10.8 & 44.6 $\pm$ 10.7 & 34.0 $\pm$ 16.8 & 40.1 $\pm$ 14.0\\
Krum         & 61.5 $\pm$ 22.1 & 7.9 $\pm$ 1.9 & 30.8 $\pm$ 11.1 & 16.2 $\pm$ 2.9 & 9.7 $\pm$ 6.5\\
BALANCE      & 22.9 $\pm$ 27.1 & 38.2 $\pm$ 35.3 & 38.2 $\pm$ 35.3 & 38.2 $\pm$ 35.3 & 38.2 $\pm$ 35.3\\
SCCLIP       & 57.9 $\pm$ 21.9 & 57.9 $\pm$ 21.9 & 57.9 $\pm$ 21.9 & 57.9 $\pm$ 21.9 & 46.6 $\pm$ 32.4\\
LIGHTYEAR    & 72.6 $\pm$ 15.6 & 75.1 $\pm$ 12.1 & 71.5 $\pm$ 12.7 & 69.2 $\pm$ 14.8 & 72.1 $\pm$ 13.6\\
\bottomrule
\end{tabular}}
\label{tab:femnist_ana_short}
\end{table*}

\begin{table*}[ht]
\centering
\caption{This table reports the results for the dynamically changing malfunction on Isic19. Average accuracy (\%) $\pm$ std of each method under an increasing number of malfunctioning clients.}
\vspace{0.5em}
\resizebox{\textwidth}{!}{%
\begin{tabular}{lccccc}
\toprule
\multicolumn{6}{c}{\textbf{Isic19 – Dyanmic Malfunctions}} \\
\midrule
\multicolumn{6}{c}{\textbf{Number of Malfunctioning Clients}} \\
\cmidrule(lr){2-6}
\textbf{Method} & \textbf{1} & \textbf{2} & \textbf{3} & \textbf{4} & \textbf{5}\\
\midrule
FedAvg       & 1.8 $\pm$ 1.4 & 57.9 $\pm$ 21.9 & 7.3 $\pm$ 8.5 & 7.3 $\pm$ 8.5 & 7.3 $\pm$ 8.5 \\
AFA          & 62.7 $\pm$ 13.6 & 62.3 $\pm$ 13.3 & 25.2 $\pm$ 10.6 & 1.6 $\pm$ 1.4  & 1.6 $\pm$ 9.6\\
ASMR         & 70.4 $\pm$ 15.5 & 59.0 $\pm$ 16.0 & 16.3 $\pm$ 9.6 & 16.3 $\pm$ 9.6 & 16.3 $\pm$ 9.6\\
CFL          & 68.9 $\pm$ 15.8 & 16.3 $\pm$ 9.6 & 48.3 $\pm$ 13.7 & 16.3 $\pm$ 9.6 & 16.3 $\pm$ 9.6\\
Ditto        & 67.1 $\pm$ 14.0 & 60.3 $\pm$ 21.1 & 57.9 $\pm$ 21.9 & 57.9 $\pm$ 21.9 & 16.3 $\pm$ 9.6\\
FedProx      & 60.7 $\pm$ 18.6 & 12.3 $\pm$ 7.6 & 12.3 $\pm$ 7.6 & 2.0 $\pm$ 2.7 & 2.0 $\pm$ 2.7\\
Krum         & 58.3 $\pm$ 22.6 & 5.5 $\pm$ 2.2 & 38.2 $\pm$ 16.5 & 16.3 $\pm$ 9.6 & 17.9 $\pm$ 4.4\\
BALANCE      & 22.9 $\pm$ 27.1 & 38.2 $\pm$ 35.3 & 38.2 $\pm$ 35.3 & 38.2 $\pm$ 35.3 & 38.2 $\pm$ 35.3\\
SCCLIP       & 9.2 $\pm$ 16.3 & 14.8 $\pm$ 7.6 & 25.7 $\pm$ 32.1 & 31.8 $\pm$ 28.4 & 6.9 $\pm$ 8.6\\
LIGHTYEAR    & 77.6 $\pm$ 12.3 & 77.6 $\pm$ 10.6 & 76.1 $\pm$ 11.5 & 74.1 $\pm$ 12.5 & 73.6 $\pm$ 9.9\\
\bottomrule
\end{tabular}}
\label{tab:femnist_ana_short}
\end{table*}

\begin{table*}[ht]
\centering
\caption{This table reports the results for the dynamically changing malfunction on the Ultrasound dataset. Average dice (\%) $\pm$ std of each method under an increasing number of malfunctioning clients.}
\vspace{0.5em}
\begin{tabular}{lcccc}
\toprule
\multicolumn{5}{c}{\textbf{Ultrasound – ANA}} \\
\midrule
\multicolumn{5}{c}{\textbf{Number of Malfunctioning Clients}} \\
\cmidrule(lr){2-5}
\textbf{Method} & \textbf{1} & \textbf{2} & \textbf{3} & \textbf{4} \\
\midrule
FedAvg       & 82.9 $\pm$ 5.4 & 52.3 $\pm$ 5.9 & 3.1 $\pm$ 0.6 & 3.1 $\pm$ 0.6 \\
AFA          & 86.1 $\pm$ 3.4 & 84.6 $\pm$ 3.2 & 65.5 $\pm$ 5.0 & 0.3 $\pm$ 0.6 \\
ASMR         & 84.3 $\pm$ 5.3 & 0.3 $\pm$ 0.6 & 0.3 $\pm$ 0.6 & 0.3 $\pm$ 0.6 \\
CFL          & 86.0 $\pm$ 3.3 & 76.5 $\pm$ 5.3 & 0.3 $\pm$ 0.6 & 0.3 $\pm$ 0.6 \\
Ditto        & 0.1 $\pm$ 0.1 & 1.0 $\pm$ 1.0 & 0.3 $\pm$ 0.6 & 0.3 $\pm$ 0.6 \\
FedProx      & 58.6 $\pm$ 6.7 & 35.1 $\pm$ 9.0 & 0.3 $\pm$ 0.6 & 0.3 $\pm$ 0.6 \\
Krum         & 79.4 $\pm$ 3.2 & 82.7 $\pm$ 1.9 & 81.1 $\pm$ 6.7 & 77.9 $\pm$ 2.0 \\
BALANCE      & 0.3 $\pm$ 0.6 & 0.3 $\pm$ 0.6 & 0.3 $\pm$ 0.6 & 0.3 $\pm$ 0.6 \\
SCCLIP       & 0.3 $\pm$ 0.6 & 0.3 $\pm$ 0.6 & 0.3 $\pm$ 0.6 & 0.3 $\pm$ 0.6 \\
LIGHTYEAR    & 83.1 $\pm$ 3.5 & 84.0 $\pm$ 3.5 & 84.2 $\pm$ 3.5 & 84.9 $\pm$ 3.5 \\
\bottomrule
\end{tabular}
\label{tab:femnist_ana_short}
\end{table*}

\begin{table*}[ht]
\centering
\caption{This table reports the results for the dynamically changing malfunction on the Ultrasound dataset. Average dice (\%) $\pm$ std of each method under an increasing number of malfunctioning clients.}
\vspace{0.5em}
\begin{tabular}{lcccc}
\toprule
\multicolumn{5}{c}{\textbf{Ultrasound – SFA}} \\
\midrule
\multicolumn{5}{c}{\textbf{Number of Malfunctioning Clients}} \\
\cmidrule(lr){2-5}
\textbf{Method} & \textbf{1} & \textbf{2} & \textbf{3} & \textbf{4} \\
\midrule
FedAvg       & 0.3 $\pm$ 0.6 & 0.3 $\pm$ 0.6 & 0.3 $\pm$ 0.6 & 9.0 $\pm$ 1.2 \\
AFA          & 85.4 $\pm$ 2.6 & 85.2 $\pm$ 2.5 & 0.3 $\pm$ 0.6 & 0.3 $\pm$ 0.6 \\
ASMR         & 85.8 $\pm$ 2.7 & 84.7 $\pm$ 3.9 & 0.3 $\pm$ 0.6 & 0.3 $\pm$ 0.6 \\
CFL          & 86.6 $\pm$ 3.5 & 83.8 $\pm$ 5.5 & 0.3 $\pm$ 0.6 & 0.3 $\pm$ 0.6 \\
Ditto        & 0.3 $\pm$ 0.6 & 0.3 $\pm$ 0.6 & 0.3 $\pm$ 0.6 & 0.3 $\pm$ 0.6 \\
FedProx      & 0.3 $\pm$ 0.6 & 0.3 $\pm$ 0.6 & 0.3 $\pm$ 0.6 & 9.0 $\pm$ 1.2 \\
Krum         & 81.4 $\pm$ 3.9 & 0.3 $\pm$ 0.6 & 0.3 $\pm$ 0.6 & 0.3 $\pm$ 0.6 \\
BALANCE      & 0.3 $\pm$ 0.6 & 0.3 $\pm$ 0.6 & 0.3 $\pm$ 0.6 & 0.3 $\pm$ 0.6 \\
SCCLIP       & 0.3 $\pm$ 0.6 & 0.3 $\pm$ 0.6 & 3.1 $\pm$ 3.8 & 5.2 $\pm$ 4.4 \\
LIGHTYEAR    & 83.6 $\pm$ 4.4 & 81.8 $\pm$ 5.1 & 82.8 $\pm$ 4.0 & 81.3 $\pm$ 4.3 \\
\bottomrule
\end{tabular}
\label{tab:femnist_ana_short}
\end{table*}

\begin{table*}[ht]
\centering
\caption{This table reports the results for the dynamically changing malfunction on the Ultrasound dataset. Average dice (\%) $\pm$ std of each method under an increasing number of malfunctioning clients.}
\vspace{0.5em}
\begin{tabular}{lcccc}
\toprule
\multicolumn{5}{c}{\textbf{Ultrasound – Random}} \\
\midrule
\multicolumn{5}{c}{\textbf{Number of Malfunctioning Clients}} \\
\cmidrule(lr){2-5}
\textbf{Method} & \textbf{1} & \textbf{2} & \textbf{3} & \textbf{4} \\
\midrule
FedAvg       & 0.4 $\pm$ 0.6 & 0.3 $\pm$ 0.6 & 0.3 $\pm$ 0.6 & 0.3 $\pm$ 0.6 \\
AFA          & 85.7 $\pm$ 3.1 & 0.3 $\pm$ 0.6 & 0.3 $\pm$ 0.6 & 0.3 $\pm$ 0.6 \\
ASMR         & 85.9 $\pm$ 2.8 & 84.4 $\pm$ 4.6 & 9.0 $\pm$ 1.2 & 9.0 $\pm$ 1.2 \\
CFL          & 85.0 $\pm$ 4.4 & 85.9 $\pm$ 2.9 & 0.3 $\pm$ 0.6 & 0.3 $\pm$ 0.6 \\
Ditto        & 0.0 $\pm$ 0.0 & 9.1 $\pm$ 1.2 & 9.0 $\pm$ 1.2 & 0.0 $\pm$ 0.0 \\
FedProx      & 0.3 $\pm$ 0.6 & 0.3 $\pm$ 0.6 & 0.3 $\pm$ 0.6 & 0.0 $\pm$ 0.0 \\
Krum         & 82.1 $\pm$ 4.9 & 0.3 $\pm$ 0.6 & 0.3 $\pm$ 0.6 & 0.3 $\pm$ 0.6 \\
BALANCE      & 0.3 $\pm$ 0.6 & 0.3 $\pm$ 0.6 & 0.3 $\pm$ 0.6 & 0.3 $\pm$ 0.6 \\
SCCLIP       & 0.3 $\pm$ 0.6 & 0.3 $\pm$ 0.6 & 0.3 $\pm$ 0.6 & 0.3 $\pm$ 0.6 \\
LIGHTYEAR    & 78.8 $\pm$ 9.8 & 81.0 $\pm$ 4.2 & 81.5 $\pm$ 4.3 & 79.1 $\pm$ 7.3 \\
\bottomrule
\end{tabular}
\label{tab:femnist_ana_short}
\end{table*}

\begin{table*}[ht]
\centering
\caption{This table reports the results for the dynamically changing malfunction on the Ultrasound dataset. Average dice (\%) $\pm$ std of each method under an increasing number of malfunctioning clients.}
\vspace{0.5em}
\begin{tabular}{lcccc}
\toprule
\multicolumn{5}{c}{\textbf{Ultrasound – Dynamic Malfunctions}} \\
\midrule
\multicolumn{5}{c}{\textbf{Number of Malfunctioning Clients}} \\
\cmidrule(lr){2-5}
\textbf{Method} & \textbf{1} & \textbf{2} & \textbf{3} & \textbf{4} \\
\midrule
FedAvg       & 0.3 $\pm$ 0.6 & 0.3 $\pm$ 0.6 & 0.3 $\pm$ 0.6 & 0.3 $\pm$ 0.6 \\
AFA          & 85.2 $\pm$ 2.6 & 85.3 $\pm$ 3.9 & 0.3 $\pm$ 0.6 & 0.0 $\pm$ 0.0 \\
ASMR         & 84.9 $\pm$ 4.7 & 0.3 $\pm$ 0.6 & 0.3 $\pm$ 0.6 & 0.3 $\pm$ 0.6 \\
CFL          & 84.5 $\pm$ 3.0 & 80.7 $\pm$ 6.5 & 0.3 $\pm$ 0.6 & 0.3 $\pm$ 0.6 \\
Ditto        & 7.4 $\pm$ 1.7 & 9.0 $\pm$ 1.2 & 0.3 $\pm$ 0.6 & 9.0 $\pm$ 1.2 \\
FedProx      & 0.3 $\pm$ 0.6 & 0.3 $\pm$ 0.6 & 0.3 $\pm$ 0.6 & 0.3 $\pm$ 0.6 \\
Krum         & 77.7 $\pm$ 7.1 & 0.3 $\pm$ 0.6 & 9.0 $\pm$ 1.2 & 70.9 $\pm$ 1.2 \\
BALANCE      & 0.3 $\pm$ 0.6 & 0.3 $\pm$ 0.6 & 0.3 $\pm$ 0.6 & 0.3 $\pm$ 0.6 \\
SCCLIP       & 0.3 $\pm$ 0.6 & 0.3 $\pm$ 0.6 & 0.3 $\pm$ 0.6 & 5.6 $\pm$ 4.7 \\
LIGHTYEAR    & 82.1 $\pm$ 4.0 & 81.9 $\pm$ 6.3 & 77.7 $\pm$ 7.5 & 78.6 $\pm$ 5.8 \\
\bottomrule
\end{tabular}
\label{tab:femnist_ana_short}
\end{table*}

\begin{table*}[ht]
\centering
\caption{This table reports the results for the dynamically changing malfunction on the XRay dataset. Average dice (\%) $\pm$ std of each method under an increasing number of malfunctioning clients.}
\vspace{0.5em}
\begin{tabular}{lcccc}
\toprule
\multicolumn{5}{c}{\textbf{XRay – ANA}} \\
\midrule
\multicolumn{5}{c}{\textbf{Number of Malfunctioning Clients}} \\
\cmidrule(lr){2-5}
\textbf{Method} & \textbf{1} & \textbf{2} & \textbf{3} & \textbf{4} \\
\midrule
FedAvg       & 79.3 $\pm$ 3.4 & 39.7 $\pm$ 4.0 & 38.8 $\pm$ 2.5 & 0.0 $\pm$ 0.0 \\
AFA          & 86.8 $\pm$ 1.9 & 85.3 $\pm$ 2.3 & 0.0 $\pm$ 0.0 & 0.0 $\pm$ 0.0 \\
ASMR         & 87.1 $\pm$ 1.3 & 80.3 $\pm$ 3.5 & 0.0 $\pm$ 0.0 & 0.0 $\pm$ 0.0 \\
CFL          & 85.8 $\pm$ 3.0 & 58.2 $\pm$ 5.9 & 0.0 $\pm$ 0.0 & 0.0 $\pm$ 0.0 \\
Ditto        & 6.6 $\pm$ 1.9 & 0.2 $\pm$ 0.2 & 0.0 $\pm$ 0. & 0.0 $\pm$ 0.0 \\
FedProx      & 80.1 $\pm$ 3.7 & 0.0 $\pm$ 0.0 & 0.0 $\pm$ 0.0 & 0.0 $\pm$ 0.0 \\
Krum         & 82.8 $\pm$ 3.5 & 85.5 $\pm$ 3.1 & 84.4 $\pm$ 3.0 & 79.2 $\pm$ 3.6 \\
BALANCE      & 39.1 $\pm$ 15.5 & 78.7 $\pm$ 3.6 & 0.0 $\pm$ 0.0 & 67.0 $\pm$ 5.1 \\
SCCLIP       & 0.0 $\pm$ 0. & 0.0 $\pm$ 0.0 & 0.0 $\pm$ 0.0 & 0.0 $\pm$ 0.0 \\
LIGHTYEAR    & 84.4 $\pm$ 1.4 & 83.8 $\pm$ 2.5 & 83.3 $\pm$ 1.9 & 83.4 $\pm$ 2.0 \\
\bottomrule
\end{tabular}
\label{tab:femnist_ana_short}
\end{table*}

\begin{table*}[ht]
\centering
\caption{This table reports the results for the dynamically changing malfunction on the XRay dataset. Average dice (\%) $\pm$ std of each method under an increasing number of malfunctioning clients.}
\vspace{0.5em}
\begin{tabular}{lcccc}
\toprule
\multicolumn{5}{c}{\textbf{XRay – SFA}} \\
\midrule
\multicolumn{5}{c}{\textbf{Number of Malfunctioning Clients}} \\
\cmidrule(lr){2-5}
\textbf{Method} & \textbf{1} & \textbf{2} & \textbf{3} & \textbf{4} \\
\midrule
FedAvg       & 0.0 $\pm$ 0.0 & 0.0 $\pm$ 0.0 & 34.0 $\pm$ 2.5 & 34.0 $\pm$ 2.5 \\
AFA          & 86.5 $\pm$ 2.0 & 86.2 $\pm$ 2.6 & 0.0 $\pm$ 0.0 & 0.0 $\pm$ 0.0 \\
ASMR         & 86.8 $\pm$ 1.8 & 85.2 $\pm$ 3.1 & 0.0 $\pm$ 0.0 & 0.0 $\pm$ 0.0 \\
CFL          & 87.0 $\pm$ 1.3 & 85.2 $\pm$ 2.9 & 0.0 $\pm$ 0.0 & 0.0 $\pm$ 0.0 \\
Ditto        & 0.0 $\pm$ 0.0 & 0.0 $\pm$ 0.0 & 0.0 $\pm$ 0.0 & 0.0 $\pm$ 0.0 \\
FedProx      & 0.0 $\pm$ 0.0 & 0.0 $\pm$ 0.0 & 34.0 $\pm$ 2.5 & 34.0 $\pm$ 2.5 \\
Krum         & 85.0 $\pm$ 1.8 & 80.3 $\pm$ 2.3 & 0.0 $\pm$ 0.0 & 0.0 $\pm$ 0.0 \\
BALANCE      & 19.0 $\pm$ 11.8 & 0.0 $\pm$ 0.0 & 0.0 $\pm$ 0.0 & 0.0 $\pm$ 0.0 \\
SCCLIP       & 0.0 $\pm$ 0.0 & 0.0 $\pm$ 0.0 & 14.2 $\pm$ 17.4 & 34.0 $\pm$ 25.3 \\
LIGHTYEAR    & 83.7 $\pm$ 1.3 & 82.9 $\pm$ 1.3 & 82.9 $\pm$ 1.1 & 81.4 $\pm$ 1.9 \\
\bottomrule
\end{tabular}
\label{tab:femnist_ana_short}
\end{table*}

\begin{table*}[ht]
\centering
\caption{This table reports the results for the dynamically changing malfunction on the XRay dataset. Average dice (\%) $\pm$ std of each method under an increasing number of malfunctioning clients.}
\vspace{0.5em}
\begin{tabular}{lcccc}
\toprule
\multicolumn{5}{c}{\textbf{XRay – Random}} \\
\midrule
\multicolumn{5}{c}{\textbf{Number of Malfunctioning Clients}} \\
\cmidrule(lr){2-5}
\textbf{Method} & \textbf{1} & \textbf{2} & \textbf{3} & \textbf{4} \\
\midrule
FedAvg       & 0.0 $\pm$ 0.0 & 0.0 $\pm$ 0.0 & 0.0 $\pm$ 0.0 & 0.0 $\pm$ 0.0 \\
AFA          & 85.7 $\pm$ 1.2 & 85.9 $\pm$ 2.1 & 0.0 $\pm$ 0.0 & 0.0 $\pm$ 0.0 \\
ASMR         & 86.6 $\pm$ 1.7 & 86.2 $\pm$ 2.7 & 34.0 $\pm$ 2.5 & 34.0 $\pm$ 2.5 \\
CFL          & 86.4 $\pm$ 1.7 & 84.9 $\pm$ 2.9 & 0.0 $\pm$ 0.0 & 0.0 $\pm$ 0.0 \\
Ditto        & 0.0 $\pm$ 0.0 & 0.0 $\pm$ 0.0 & 0.0 $\pm$ 0.0 & 0.0 $\pm$ 0.0 \\
FedProx      & 0.0 $\pm$ 0.0 & 0.0 $\pm$ 0.0 & 0.0 $\pm$ 0.0 & 0.0 $\pm$ 0.0 \\
Krum         & 82.6 $\pm$ 3.9 & 85.2 $\pm$ 2.6 & 84.1 $\pm$ 3.6 & 0.0 $\pm$ 0.0 \\
BALANCE      & 6.4 $\pm$ 11.3 & 0.0 $\pm$ 0.0 & 0.0 $\pm$ 0.0 & 0.0 $\pm$ 0.0 \\
SCCLIP       & 0.0 $\pm$ 0.0 & 0.0 $\pm$ 0.0 & 5.9 $\pm$ 11.8 & 34.0 $\pm$ 15.3 \\
LIGHTYEAR    & 83.2 $\pm$ 2.0 & 82.1 $\pm$ 2.2 & 80.0 $\pm$ 3.2 & 80.3 $\pm$ 3.7 \\
\bottomrule
\end{tabular}
\label{tab:femnist_ana_short}
\end{table*}

\begin{table*}[ht]
\centering
\caption{This table reports the results for the dynamically changing malfunction on the XRay dataset. Average dice (\%) $\pm$ std of each method under an increasing number of malfunctioning clients.}
\vspace{0.5em}
\begin{tabular}{lcccc}
\toprule
\multicolumn{5}{c}{\textbf{XRay – Dynamic Malfunctions}} \\
\midrule
\multicolumn{5}{c}{\textbf{Number of Malfunctioning Clients}} \\
\cmidrule(lr){2-5}
\textbf{Method} & \textbf{1} & \textbf{2} & \textbf{3} & \textbf{4} \\
\midrule
FedAvg       & 80.9 $\pm$ 3.5 & 0.0 $\pm$ 0.0 & 0.0 $\pm$ 0.0 & 34.0 $\pm$ 2.5 \\
AFA          & 87.4 $\pm$ 1.0 & 85.3 $\pm$ 2.4 & 0.0 $\pm$ 0.0 & 0.0 $\pm$ 0.0 \\
ASMR         & 86.6 $\pm$ 1.5 & 81.9 $\pm$ 3.0 & 0.0 $\pm$ 0.0 & 0.0 $\pm$ 0.0 \\
CFL          & 86.4 $\pm$ 2.4 & 82.5 $\pm$ 2.8 & 0.0 $\pm$ 0.0 & 0.0 $\pm$ 0.0 \\
Ditto        & 0.0 $\pm$ 0.0 & 0.0 $\pm$ 0.0 & 34.0 $\pm$ 2.5 & 34.0 $\pm$ 2.5 \\
FedProx      & 0.0 $\pm$ 0.0 & 0.0 $\pm$ 0.0 & 0.0 $\pm$ 0.0 & 0.0 $\pm$ 0.0 \\
Krum         & 84.5 $\pm$ 2.8 & 84.7 $\pm$ 2.5 & 80.2 $\pm$ 3.0 & 0.0 $\pm$ 0.0 \\
BALANCE      & 27.9 $\pm$ 15.7 & 0.7 $\pm$ 1.3 & 0.0 $\pm$ 0.0 & 0.0 $\pm$ 0.0 \\
SCCLIP       & 0.0 $\pm$ 0.0 & 0.0 $\pm$ 0.0 & 0.0 $\pm$ 0.0 & 34.0 $\pm$ 2.5 \\
LIGHTYEAR    & 82.7 $\pm$ 1.5 & 82.8 $\pm$ 1.7 & 82.6 $\pm$ 3.1 & 81.9 $\pm$ 2.2 \\
\bottomrule
\end{tabular}
\label{tab:femnist_ana_short}
\end{table*}
\section{Technical Details}
In this section, we provide an overview of the hardware and software requirements needed to reproduce our experiments. We detail the dataset construction process for each client in our FL setup, ensuring reproducibility of the data splits and assignments. Additionally, we outline the model architectures used in our experiments, including all relevant hyperparameters.

\subsection{Hard- and Software Requirements}

Our implementation is based on NVIDIA FLARE, as described in the main paper. The required software configuration includes Python 3.10.17, CUDA 12.2, and PyTorch 2.7.0+cu126. We also used a nightly build of FLARE 2.6.1. All remaining dependencies are listed in the requirements.txt file provided in our codebase. Experiments were conducted on an SLURM-managed cluster using NVIDIA A100 GPUs with 40GB memory. We used up to three GPUs per training run. Each experiment can be executed with a single GPU, though. 
\subsection{Training Details} \label{apx:train}
To complement the description of our training routines provided in the main paper, we present the detailed hyperparameters for our experimental setup in Table \ref{tab:hp}. Additionally, our codebase includes the full implementation of these training routines, ensuring that all experiments are fully reproducible.

\begin{table}[ht]
\centering
\caption{This table lists the hyperparameters used for the experiments in the main paper to reproduce the results.}
\vspace{0.5em}
\resizebox{\textwidth}{!}{%
\begin{tabular}{lccccc}
\toprule
\multicolumn{6}{c}{\textbf{Training Details}} \\
\midrule
\multicolumn{6}{c}{\textbf{Datasets}} \\
\cmidrule(lr){2-6}
\textbf{Method} & \textbf{FEMNIST} & \textbf{Camelyon17} & \textbf{Isic19} & \textbf{Ultrasound} & \textbf{XRay}\\
\midrule
Loss Function & Cross Entropy & Cross Entropy  & Weighted Focal Loss & Cross Entropy & Cross Entropy\\
Optimizer     & Adam          & SGD            & Adam          & Adam          & Adam   \\
Learning Rate & 1e-3          & 1e-3           & 5e-4          & 1e-3          & 1e-3   \\
Weight Decay  & 1e-4          & 5e-4           & 0             & 0             & 0      \\
Momentum      & -             & 0.9            & -             & -             & -     \\
Batchsize     & 32            & 32             & 32            & 8             & 4     \\
Rounds        & 12            & 12             & 20            & 60            & 60   \\
Local Epochs  & 1             & 1              & 1             & 1             & 1   \\
\bottomrule
\end{tabular}}
\label{tab:hp}
\end{table}

\section{Model Sensitivity}

\begin{figure}
    \centering
    \includegraphics[width=\linewidth]{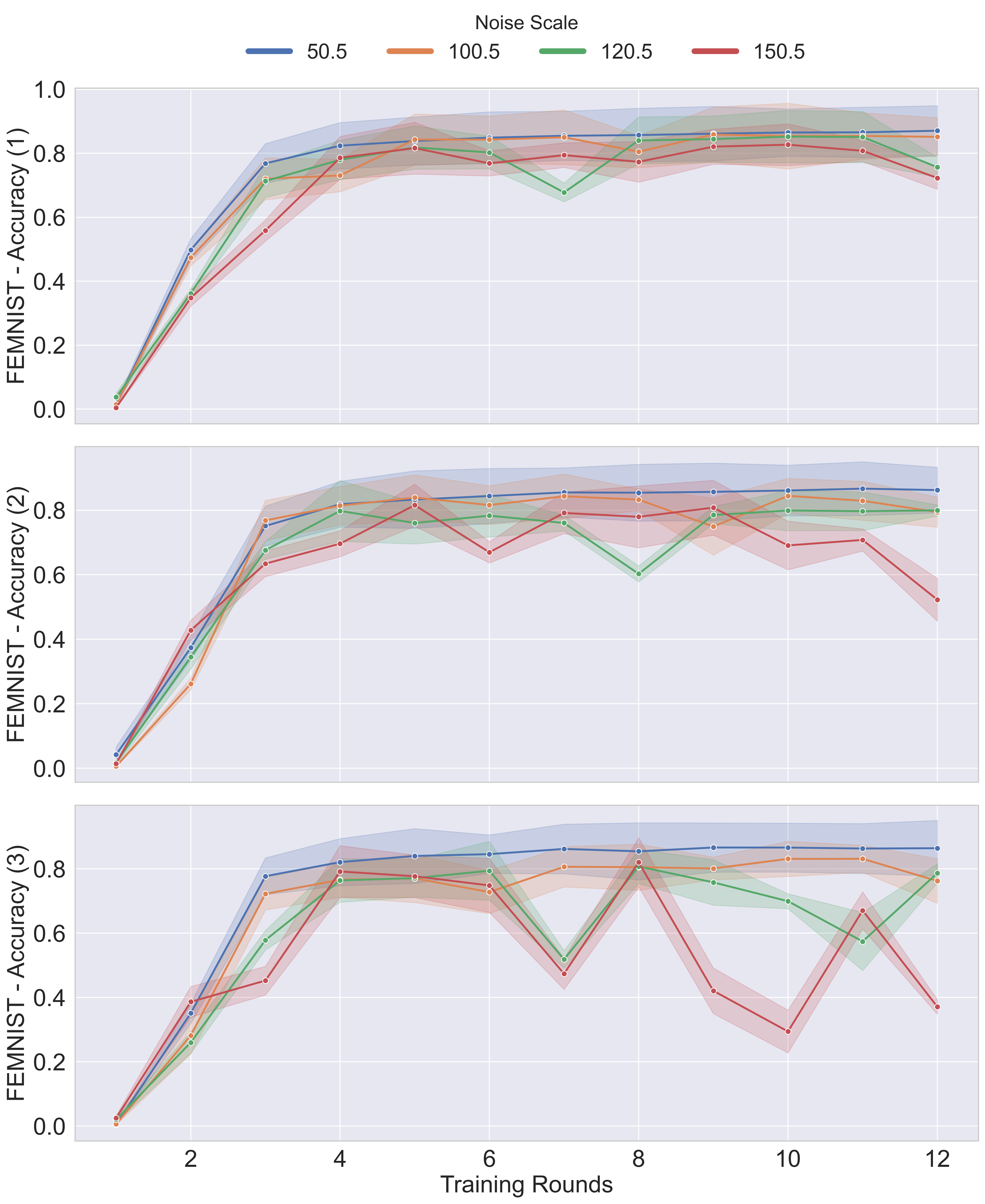}
    \caption{Training performance on FEMNIST under varying levels of corruption. Each plot corresponds to a different number of malfunctioning clients, indicated in parentheses in the y-axis label. Results show the average validation accuracy across all clients at each training round.}
    \label{fig:femnist_sensitivity}
\end{figure}

\begin{figure}
    \centering
    \includegraphics[width=\linewidth]{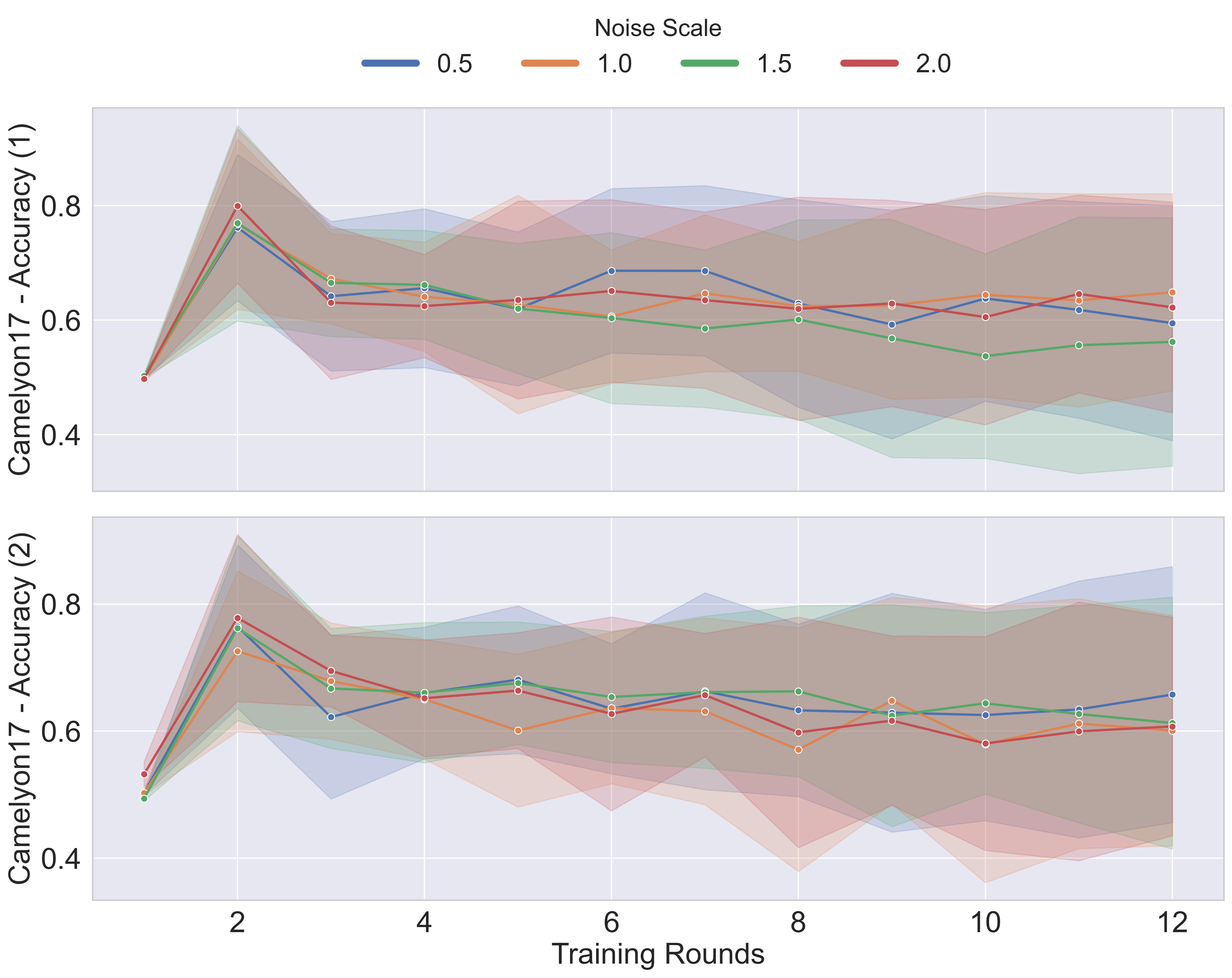}
    \caption{Training performance on Camelyon17 under varying levels of corruption. Each plot corresponds to a different number of malfunctioning clients, indicated in parentheses in the y-axis label. Results show the average validation accuracy across all clients at each training round.}
    \label{fig:camelyon17_sensitivity}
\end{figure}

To investigate the sensitivity of the models used in our study to corrupted updates, we conducted an ablation study using the ANA. In the main paper, we employed two model architectures: a DenseNet121 for the Camelyon17 dataset, chosen due to its prevalence on the official leaderboard, and a lightweight CNN with two convolutional layers for the FEMNIST dataset, which has shown strong performance on this task. To evaluate how these models react to corrupted updates, we applied ANA as defined in the main paper. In our implementation, noise is injected into the updates using a scaling factor $s$, as described in Equation \ref{eq:ana}, allowing us to vary the corruption severity from very light to heavy flexibly. This tunable corruption is an advantage over SFA, which inherently applies only strong corruption. For our ablation, we applied different scaling factors and varied the number of malfunctioning clients, ranging from 1 to 3 for FEMNIST, and from 1 to 2 for Camelyon17. We monitored how the corruption affected the learning dynamics by evaluating each client's local validation accuracy at every training round. The mean validation accuracy across all clients per round is reported in Figure \ref{fig:femnist_sensitivity} for FEMNIST and Figure \ref{fig:camelyon17_sensitivity} for Camelyon17.

\begin{equation}\label{eq:ana}
    \tilde{\theta}_k = \theta_k + \left( \epsilon_k \cdot \frac{s}{100} \cdot \theta_k \right), \quad \text{where} \quad \epsilon_k \sim \mathcal{N}(0, I)
\end{equation}

Our ablation study reveals that the model trained on the FEMNIST dataset exhibits strong robustness to noise and is therefore less sensitive to corruption. We evaluated this sensitivity using four different noise scaling factors, as detailed in Figure 1. The results indicate that even scaling factors around 50 do not destabilize the training process. However, when scaling reaches around 120, the training begins to show instability, even with only a single malfunctioning client. At a scaling factor of approximately 150, we observe significant performance degradation, particularly when three clients are malfunctioning. In contrast, the model trained on the Camelyon17 dataset is much more sensitive to noise. As a result, we evaluated this setting using substantially smaller scaling factors. Even at a scaling of 1.5, the training becomes noticeably destabilized. It is worth noting that training on Camelyon17 is inherently less stable and more challenging due to the high degree of client heterogeneity. This instability is also reflected in the main paper’s ablation study and motivated our introduction of the regularized aggregation approach. In our final experiments, we used a scaling factor of 120.5 for FEMNIST and 1.5 for Camelyon17. While these settings do not lead to the most extreme performance drops, they effectively destabilize training and negatively impact performance, striking a balance between subtle corruption and the more aggressive, rapidly diverging effect of SFAs.

\end{document}